%% file: main.tex
\documentclass[a4paper,fleqn]{cas-sc}

\usepackage[numbers]{natbib}
\def\tsc#1{\csdef{#1}{\textsc{\lowercase{#1}}\xspace}}
\tsc{WGM}
\tsc{QE}

\begin{document}
\let\WriteBookmarks\relax
\renewcommand{\topfraction}{0.95}
\renewcommand{\bottomfraction}{0.8}
\renewcommand{\textfraction}{0.05}
\renewcommand{\floatpagefraction}{0.8}

\setcounter{topnumber}{3}
\setcounter{bottomnumber}{2}
\setcounter{totalnumber}{5}
\shorttitle{}    

\shortauthors{}  

\title [mode = title]{Adaptive Dual-Constrained Line Aggregation for Cross-Paradigm Line Segment Detection}  



%

\author[1]{Chenguang Liu}



\ead{cg.l.x@outlook.com}


\credit{Conceptualization, Methodology, Software, Roles/Writing - original draft}

\author[1]{Chisheng Wang}

\cormark[1]


\ead{wangchisheng@szu.edu.cn}


\credit{Conceptualization, Methodology, Resources, Supervision, Funding acquisition, Writing - review $\&$ editing}

\author[1]{Huilin Chen}
\credit{Methodology, Software, Visualization}
\author[1]{Chuanhua Zhu}
\credit{Validation, Writing - review $\&$ editing, Funding acquisition}
\author[1]{Qingquan Li}
\credit{Supervision, Writing - review $\&$ editing, Resources, Funding acquisition}

\affiliation[1]{organization={Ministry of Natural Resources (MNR) Key Laboratory for Geo-Environmental Monitoring of Great Bay Area $\&$ Guangdong Key Laboratory of Urban Informatics $\&$ Shenzhen Key Laboratory of Spatial Smart Sensing and Services, School of Architecture $\&$ Urban Planning, Shenzhen University},
            addressline={3688 Nanhai Avenue}, 
            city={Shenzhen},
            postcode={518060}, 
            state={Guangdong Province},
            country={China}}

\cortext[1]{Corresponding author}



\begin{abstract}
Line segment detection has been studied for decades, 
yet existing methods are typically designed for different detection paradigms. 
Generic line segment detectors aim to recover all meaningful line segments in an image, 
whereas recent deep-learning-based approaches mainly target wireframe line segments that describe salient geometric structures. 
Because these paradigms follow different detection objectives, 
methods optimized for one often perform poorly on the other. 
In this work, 
we propose Adaptive Dual-Constrained Line Aggregation (ADLA), 
a line extraction framework designed to operate across different line segment detection paradigms. 
Starting from an edge strength map, 
ADLA progressively aggregates pixels into candidate line segments under two complementary geometric constraints: 
orientation coherence and bounded orthogonal distance to an adaptively estimated line model. 
During aggregation, 
the line centroid and orientation are dynamically updated using the accumulated supporting pixels, 
progressively improving the geometric consistency of the estimated line. 
Edge strength information is further incorporated into orientation estimation, seed selection, model refinement, and segment validation, 
reducing the need for extensive parameter tuning. 
Experiments on three publicly available datasets covering generic, wireframe, and Manhattan line segment detection demonstrate consistently strong performance across substantially different annotation settings. 
ADLA achieves ($\mathrm{F}^{\mathrm{H}}$) scores of 0.8665 on YorkUrban-LineSegment dataset, 0.8720 on ShanghaiTech dataset, and 0.7297 on YorkUrban dataset. 
These results demonstrate the effectiveness and flexibility of ADLA across different line segment detection paradigms.
\end{abstract}




\maketitle

\input{intro_revised}
\input{related_work_revised}
\input{adla_revised}
\input{experiments_revised}
\input{conclusions_revised}
\section*{Code availability}
The source code for this work is publicly available at
\url{https://github.com/ChenguangTelecom/adla}.

\section*{Funding}
This research was funded by the Yunnan Provincial Key Research and Development Program Project (202603AG380001), 
the National Natural Science Foundation of China (42374018; 42304012; W2521147), 
the Guangdong Basic and Applied Basic Research Foundation (2025B1515020092),
and the Intelligent Computing Center of Shenzhen University.

\section*{Declaration of generative AI and AI-assisted technologies in the manuscript preparation process}
During the preparation of this work, 
the author(s) used ChatGPT (OpenAI) to improve the language and readability of the manuscript. 
After using this tool/service, the author(s) reviewed and edited the content as needed and take(s) full responsibility for the content of the published article.

\printcredits

\bibliographystyle{elsarticle-num} 

\bibliography{all_ref}



\end{document}

%% file: intro_revised.tex
\section{Introduction}
\label{intro}
Line segments are fundamental image primitives that compactly describe man-made structures such as buildings, roads, and indoor layouts. 
They support many computer-vision tasks, 
including vanishing-point estimation \cite{vp2014}, 3D reconstruction \cite{3D_line}, image segmentation \cite{gpb}, object detection \cite{object_line}, and stereo matching \cite{stereo_line}.

Classical detectors operate on gradients or edge maps and recover segments through the Hough transform \cite{hough72,hough87} 
or perceptual grouping \cite{burns86,lsd}. 
Generic detectors such as LSD \cite{lsd} and Linelet \cite{linelet} seek all meaningful segments supported by image evidence, 
using validation criteria to control false detections \cite{lsd,meaningful00,linelet}. 
We call this setting \emph{generic line segment detection}.

Wireframe detection has a different objective: 
it retains segments describing salient geometry in man-made scenes 
while annotations may exclude shadows, textures, and irregular objects \cite{dwp}. 
The ShanghaiTech dataset \cite{dwp} enabled learning-based methods that jointly reason about junctions and segments \cite{lcnn,ppgnet}, 
generate proposals from attraction fields \cite{afm,afmjournal,hawp,hawpv2}, 
use alternative parameterizations \cite{tplsd,elsd}, 
or employ transformers \cite{dtlsd,linea,rankletr}.

Although both settings are called line segment detection, 
their objectives differ. 
A generic detector may be penalized on a wireframe benchmark for valid but unlabeled non-wireframe segments, 
whereas a wireframe detector may miss many segments under exhaustive generic annotations. 
YorkUrban dataset \cite{3D_line} introduces another setting by retaining only Manhattan-world segments. 
Consequently, 
rankings on ShanghaiTech dataset or YorkUrban dataset do not necessarily reflect generic detection capability. 
Yet recent wireframe detectors have not been systematically evaluated on YorkUrban-LineSegment dataset \cite{linelet}, 
which uses the same YorkUrban images but annotates all meaningful segments. 
Such evaluation reveals whether progress under structural annotations transfers to generic detection.

Training data also differ across paradigms. 
ShanghaiTech dataset supplies enough annotations for supervised wireframe learning, 
but comparably large datasets are unavailable for generic or Manhattan detection. 
This asymmetry motivates separating task-appropriate edge evidence from the geometric conversion of that evidence into segments. 
A robust aggregation stage could then process different edge-strength maps (ESMs) without task-specific line-segment training data.

We therefore propose Adaptive Dual-Constrained Line Aggregation (ADLA), 
a geometric line extractor operating on ESMs. 
ADLA estimates the orientation of each nonzero pixel from local directional responses 
and progressively groups pixels under two complementary constraints: 
orientation coherence and bounded orthogonal distance to an adaptive line model. 
As support accumulates, 
it updates the line centroid and orientation, 
improving geometric consistency. 
Edge strength also guides seed selection, orientation estimation, model refinement, and validation. 
A minimum-support criterion derived from meaningful alignments \cite{meaningful00} determines final acceptance.

Because ADLA does not encode a task-specific definition of valid segments, 
it can be paired with task-appropriate ESMs for generic, wireframe, and Manhattan annotations. 
This design also permits direct cross-paradigm comparison. 
Our results show that baseline rankings change substantially with the annotation objective, 
whereas ADLA performs consistently across all three settings.

Our contributions are:
\begin{itemize}
    \item a cross-paradigm study showing how generic, wireframe, and Manhattan annotations affect comparisons, 
    including a systematic evaluation of representative recent wireframe detectors on YorkUrban-LineSegment dataset;
    \item ADLA, which combines orientation and orthogonal-distance constraints with progressive, edge-strength-weighted refinement, 
    without task-specific line-segment supervision; and
    \item experiments on three public benchmarks demonstrating strong, consistent performance with limited adjustment of aggregation parameters.
\end{itemize}

Section \ref{related_work} reviews related methods, 
Section \ref{adla} presents ADLA, 
Section \ref{experiments} reports the experiments, 
and Section \ref{conclusions} concludes the paper.

%% file: related_work_revised.tex
\section{Related work}
\label{related_work}

Existing line detectors pursue different objectives. 
Classical methods generally recover all meaningful image-supported segments, 
whereas recent learning-based methods mainly parse structurally salient wireframes. 
We review both groups and their cross-paradigm relationship.

\subsection{Generic line segment detectors}

Early methods commonly used the Hough transform \cite{hough72,hough87} after an edge detector such as Canny \cite{canny}. 
By searching parameter space for collinear support, 
they convert high-response line hypotheses into segments. 
However, 
dense edges may cause false detections \cite{lsd,linelet}, 
digitization may produce multiple responses for one segment and performance can require careful parameter selection \cite{lsdsar}; 
numerous extensions address these limitations \cite{ppht,hough2011,hough2013}.

Perceptual-grouping methods instead combine locally consistent pixels. 
Burns et al. \cite{burns86} grouped pixels by gradient orientation. 
LSD \cite{lsd} added efficient detection and an \textit{a contrario} validation criterion based on the Helmholtz principle \cite{meaningful00}, 
controlling accidental detections with little tuning, 
although strict validation can reject weak or complex true segments \cite{linelet}. 
Linelet \cite{linelet} models digitization-induced connectivity 
and combines linelet aggregation with probabilistic validation to improve recovery under generic annotations.

Recent hybrid method combines learned representations with geometric extraction. 
DeepLSD \cite{deeplsd} predicts a line-attraction field, 
converts it to surrogate gradient maps, 
and applies classical detection. 
AAGLSD \cite{aaglsd} uses aligned anchor groups for line prediction. 

\subsection{Wireframe line segment detectors}

ShanghaiTech dataset \cite{dwp} annotates segments associated with salient scene geometry 
while omitting many lines caused by shadows, textures, or irregular objects. 
Wireframe detectors therefore recover a structural subset rather than every image-supported line.

Early learned methods decomposed detection into junction localization, proposal generation, and verification. 
DWP \cite{dwp} predicts junctions and segments separately; 
L-CNN \cite{lcnn} connects junction proposals and verifies the candidates. 
HAWP \cite{hawp} uses attraction fields to generate proposals, 
and HAWPv2 \cite{hawpv2} improves junction--line binding and verification. 
PPGNet \cite{ppgnet} and LGNN \cite{lgnn} model proposal relationships as graphs.

Other methods introduce alternative representations. 
AFM \cite{afm} recovers segments from an attraction field; 
TP-LSD \cite{tplsd} uses a root point and two endpoints; 
ELSD \cite{elsd} and F-Clip \cite{fclip} use center--length--orientation parameterizations. 
LSDNet \cite{lsdnet}, M-LSD \cite{mlsd}, and EM-LSD \cite{emlsd} emphasize efficiency, 
while Deep Hough formulations inject geometric priors through trainable Hough components \cite{deephough}.

Transformers have also become prominent. 
LETR \cite{letr} adapts DETR \cite{detr} to direct endpoint regression, 
RANK-LETR \cite{rankletr} improves training and ranking, 
DT-LSD \cite{dtlsd} uses deformable attention, 
and LINEA \cite{linea} provides a scalable architecture. 
To reduce reliance on manual labels, 
HAWPv3 \cite{hawpv2} uses self-supervision and ScaleLSD \cite{scalelsd} learns from more than $10$ million unlabeled images.

\subsection{Cross-paradigm perspective}

Generic detectors preserve all meaningful image-supported segments; 
wireframe detectors retain salient structural subsets; 
and YorkUrban dataset \cite{3D_line} retains only Manhattan-world segments. 
Thus, 
a valid generic segment may be counted as a false positive under wireframe or Manhattan annotations, 
while structurally specialized detectors may have low recall against exhaustive annotations. 
Strong performance on ShanghaiTech dataset or YorkUrban dataset therefore does not guarantee generic detection ability, 
and generic methods may be unfairly penalized on structural benchmarks.

This gap is amplified by training data: 
ShanghaiTech dataset supports large-scale supervised learning, 
whereas exhaustive generic annotations remain scarce. 
We consequently separate task-appropriate edge evidence from a shared geometric aggregation process. 
ADLA aggregates suitable ESMs using orientation-coherence and orthogonal-distance constraints with adaptive model refinement, 
allowing the same extractor to be tested across generic, wireframe, and Manhattan paradigms.

%% file: adla_revised.tex
\section{Adaptive Dual-Constrained Line Aggregation}
\label{adla}

The proposed framework separates task-dependent edge evidence from a task-independent geometric line aggregation procedure. 
This separation is central to the cross-paradigm setting considered in this work: 
generic, wireframe, and Manhattan line segment detection differ in the type of line structures that should be preserved, 
whereas the geometric process of grouping aligned edge evidence into line segments can remain shared across these paradigms.
Accordingly, the framework consists of three stages: 
(1) computation of a task-appropriate edge strength map (ESM), 
(2) estimation of local edge orientations from the ESM, and 
(3) Adaptive Dual-Constrained Line Aggregation (ADLA), which progressively groups supporting pixels into candidate line segments.
The first stage adapts the input evidence to the target detection paradigm, 
while the latter two stages provide a common geometric line-extraction mechanism.
The three stages are described below.

\subsection{Task-appropriate edge strength map computation}
\label{esm_computation}

The role of the first stage is to provide edge evidence that is consistent with the annotation objective of the target task.
For generic line segment detection, 
the ESM should preserve all meaningful image edges that may support valid line segments.
For wireframe detection, 
it should instead emphasize edges associated with salient geometric structures while suppressing edges that are not part of the wireframe annotations.
For Manhattan line segment detection, 
the same line aggregation mechanism can be applied to the available task-appropriate edge evidence, 
while the benchmark annotations determine which detected structures are regarded as valid during evaluation.
This formulation allows the subsequent ADLA procedure to remain unchanged across different line segment detection paradigms.

In this work, 
EdgeNAT \cite{edgenat}, a transformer-based edge detector built on the Dilated Neighborhood Attention Transformer (DiNAT) \cite{dinat}, is used to generate the ESMs.
For generic line segment detection, 
EdgeNAT is trained on BIPEDv2 dataset \cite{dexined}, 
whose annotations are designed to preserve meaningful image boundaries.
The data augmentation strategy follows \cite{dexined}.
For wireframe and Manhattan line segment detection, 
EdgeNAT is trained using the training split of the ShanghaiTech dataset.
The annotated wireframe line segments are rasterized into binary edge maps: 
a pixel is assigned a value of 1 if it belongs to an annotated line segment and 0 otherwise.
The training images and their corresponding wireframe edge maps are augmented by horizontal flipping, vertical flipping, and simultaneous horizontal and vertical flipping, and are rescaled to $50\%$, $80\%$, $100\%$, $120\%$, and $150\%$ of their original size.

We use a PyTorch implementation of EdgeNAT 
and optimize it with AdamW optimizer using a weighted cross-entropy loss.
The network is initialized from DiNAT weights pre-trained on ImageNet-22K \cite{imagenet}.
The learning rate, mini-batch size, and weight decay are set to $6\times10^{-5}$, $8$, and $0.01$, respectively.
The models are trained for 20 epochs, and the learning rate is decreased by 10 after 12 epochs.
For generic edge detection, 
the checkpoint attaining the highest edge-detection accuracy on the BIPEDv2 test set is used to generate the ESMs evaluated on YorkUrban-LineSegment dataset. 
For ShanghaiTech and YorkUrban dataset, 
we use the final checkpoint obtained after 20 training epochs.

During inference, the multi-scale testing strategy of \cite{edgenat} is used to improve edge detection accuracy.
Standard non-maximum suppression \cite{structurededge} is then applied to obtain a thinned ESM.
The resulting ESM serves as the input to the orientation-estimation and line-aggregation stages.
Importantly, once the ESM has been generated, 
the subsequent geometric processing is identical across the detection paradigms considered in this work.

\subsection{ESM-based orientation estimation}
\label{orientation_computation}

ADLA requires a local orientation estimate for each nonzero pixel in the ESM.
Rather than computing orientations from image gradients directly, 
we estimate them from the spatial distribution of edge strengths.
This makes the orientation estimation consistent with the learned or hand-crafted edge evidence supplied to the framework.

For each pixel $u(x,y)>0$ in an ESM $u$, 
responses are evaluated along $P$ uniformly sampled directions in $[0,\pi)$.
The orientation associated with the maximum directional response is assigned to the pixel.

For the $i$-th direction, 
$i=1,2,\ldots,P$, 
the corresponding angle is
\begin{align}
\theta_i = \frac{(i-1)\pi}{P}.
\end{align}
The directional response at $(x,y)$ is computed as
\begin{align}
S_i(x,y)
=
\sum_{x'=-W}^{W}
\sum_{y'=-W}^{W}
u(x+x',y+y')\,Wt_i(x',y'),
\end{align}
where $Wt_i(\cdot)$ is a line-shaped window aligned with $\theta_i$.
For an offset $(x',y')$, its value is defined as
\begin{align}
Wt_i(x',y')= \begin{cases}
1.0 & \text{ if $|-x'*\sin (\theta_i)+y'*\cos(\theta_i)|$} \\
    & \text{ $<0.5$ and $x'^2+y'^2\le W^2$,} \\
0.0 & \text{otherwise.}\\
\end{cases}
\end{align}

The orientation assigned to $(x,y)$ is therefore
\begin{align}
\theta(x,y)=\theta_{i^\ast},
\qquad
i^\ast=\arg\max_i S_i(x,y).
\end{align}
In all experiments, 
$W$ is set to 7 and $P$ to 16.
The resulting orientation map provides one of the two geometric constraints used during line aggregation.

\subsection{Adaptive Dual-Constrained Line Aggregation}
\label{region_grow}

Given an ESM and its orientation map, 
ADLA progressively aggregates edge pixels into candidate line segments.
Its central design is the simultaneous enforcement of two complementary geometric constraints:
\textit{orientation coherence} and \textit{bounded orthogonal distance}.
The first encourages pixels belonging to the same candidate segment to have compatible local orientations, 
while the second restricts them to a narrow spatial support around the current line model.
Together, 
the two constraints reduce the risk of grouping unrelated but nearby edge responses.

Let a seed pixel initiate a candidate line segment.
A neighboring pixel can be incorporated only if
(1) its orientation differs from that of the seed by no more than an angular tolerance $\tau$, and
(2) its orthogonal distance to the current line model does not exceed $l_w$ pixels.
Suppose that the current line model is specified by a reference point $(x_0,y_0)$ 
and an orientation $\theta_0$.
The orthogonal distance from a pixel $(x_1,y_1)$ to this line is
\begin{align}
d_{\perp}
=
\left|
-(x_1-x_0)\sin(\theta_0)
+
(y_1-y_0)\cos(\theta_0)
\right|.
\end{align}

A fixed line model would make the aggregation sensitive to errors in the initial seed orientation and position.
ADLA therefore updates both the reference point and the line orientation as additional supporting pixels are accumulated.
This adaptive refinement is a key distinction of the proposed aggregation strategy:
the geometric model is progressively re-estimated from the evidence already assigned to the candidate segment, 
rather than remaining fixed throughout line aggregation.
As a result, 
the line hypothesis can become increasingly consistent with the accumulated edge support.

\subsubsection{Seed selection and confidence-guided processing}

The reliability of the seed is important 
because the orientation-coherence constraint is defined relative to its orientation.
To reduce the influence of weak or noisy edge responses, 
only pixels whose ESM value exceeds a threshold $\lambda$ are allowed to initialize a candidate segment.
Pixels with lower edge strength cannot act as seeds, but they may still be incorporated into an existing segment if they satisfy both geometric constraints.
This allows weak but geometrically consistent evidence to contribute without allowing such responses to initiate potentially unstable hypotheses.

The ESM values are divided into ten equal-width intervals,
$(0,0.1], (0.1,0.2],\ldots,(0.9,1.0]$.
The bins are processed in descending order of edge strength.
Consequently, 
candidate segments are preferentially initialized from stronger edge evidence, 
while weaker responses are considered later.
Edge strength is therefore used not only to represent the likelihood of an edge, 
but also to guide the order in which geometric hypotheses are formed.

\subsubsection{Dual-constrained aggregation}

Starting from a seed pixel, ADLA examines its $s\times s$ neighborhood.
A pixel is added to the current support set if its orientation is compatible with the seed orientation within $\tau$ and its orthogonal distance to the current line model is no greater than $l_w$.
After a new pixel is accepted, the $s\times s$ neighborhoods of all aggregated pixels are recursively examined.
The procedure continues until no additional pixel satisfies both constraints.

Each accepted pixel is temporarily marked as \textit{USED} so that it is not processed repeatedly during the construction of the same hypothesis.
When aggregation terminates, the resulting support set is validated.
If its effective support is at least $l_{\min}$, the set is accepted as supporting a meaningful line segment and the segment endpoints are estimated.
Otherwise, the hypothesis is rejected and its pixels are reset to \textit{UNUSED}, allowing them to participate in subsequent hypotheses.
The procedure then proceeds to the next eligible seed and terminates when all possible seed pixels have been examined.

This aggregation process depends only on the geometry and confidence values contained in the ESM.
It does not encode a task-specific definition of a generic, wireframe, or Manhattan line segment.
The task dependence is instead introduced through the edge evidence supplied to ADLA, 
which is what enables the same aggregation mechanism to be used in the cross-paradigm experiments.

\subsubsection{Adaptive refinement of the line model}

At initialization, 
the seed pixel serves as the reference point and its estimated orientation defines the initial line direction.
As more pixels are aggregated, 
the reference point is updated to the edge-strength-weighted centroid of the current support set, 
and the line direction is re-estimated from its weighted spatial distribution.

Let $(x(i),y(i))$ denote the coordinate of the $i$-th pixel in a support set containing $N_s$ pixels, 
and let $p(x(i),y(i))$ denote its ESM value.
The weighted centroid $(c_x,c_y)$ is
\begin{align}
c_x
=
\frac{\sum_{i=1}^{N_s}p(x(i),y(i))x(i)}
{\sum_{i=1}^{N_s}p(x(i),y(i))},
\end{align}
and
\begin{align}
c_y
=
\frac{\sum_{i=1}^{N_s}p(x(i),y(i))y(i)}
{\sum_{i=1}^{N_s}p(x(i),y(i))}.
\end{align}
The line orientation is obtained from the major principal axis of the weighted scatter matrix
\begin{align}
\mathbb{M}
=
\left(
\begin{array}{cc}
\mathbb{M}^{xx} & \mathbb{M}^{xy}\\
\mathbb{M}^{xy} & \mathbb{M}^{yy}
\end{array}
\right),
\end{align}
where
\begin{align}
\mathbb{M}^{xx}
&=
\frac{\sum_{i=1}^{N_s}p(x(i),y(i))(x(i)-c_x)^2}
{\sum_{i=1}^{N_s}p(x(i),y(i))}, \nonumber\\
\mathbb{M}^{yy}
&=
\frac{\sum_{i=1}^{N_s}p(x(i),y(i))(y(i)-c_y)^2}
{\sum_{i=1}^{N_s}p(x(i),y(i))}, \nonumber\\
\mathbb{M}^{xy}
&=
\frac{\sum_{i=1}^{N_s}p(x(i),y(i))(x(i)-c_x)(y(i)-c_y)}
{\sum_{i=1}^{N_s}p(x(i),y(i))}. \nonumber
\end{align}
The updated line orientation is the angle of the eigenvector associated with the largest eigenvalue of $\mathbb{M}$.
Weighting the centroid and scatter matrix by the ESM values gives stronger edge evidence a larger influence on the refined geometric model.

Re-estimating the line model after every newly aggregated pixel would be unnecessary 
and could make the early estimates overly sensitive to small support sets.
ADLA therefore updates the reference point and orientation only when the distance between the newly aggregated pixel and the current reference point exceeds an adaptive threshold $dist_r$.

To determine this threshold, 
consider the maximum angular discrepancy induced by orientation quantization 
and the permitted aggregation tolerance.
Because $P$ orientations are uniformly sampled over $[0,\pi)$, 
the quantization error is at most $\pi/(2P)$.
In addition, 
pixels whose orientation differs from the seed by at most $\tau$ may be incorporated.
With $\tau=\pi/P$, 
the maximum discrepancy considered here is therefore
$3\pi/(2P)$.
Given the permitted orthogonal deviation $l_w$, 
the corresponding longitudinal distance is
\begin{align}
\frac{l_w}{\sin\left(\frac{3\pi}{2P}\right)}.
\end{align}
Accordingly, 
the update threshold is defined as
\begin{align}
dist_r
=
idx
\frac{l_w}{\sin\left(\frac{3\pi}{2P}\right)},
\end{align}
where $idx$ is initialized to 1 and incremented after each model update.
This strategy postpones successive re-estimations 
until the support set extends farther along the candidate segment, 
so that later updates are based on increasingly broad spatial evidence.

\subsubsection{Meaningful-support validation and endpoint estimation}

After aggregation, 
a candidate must contain sufficient support to be accepted as a line segment.
We derive the minimum support threshold $l_{\min}$ from the notion of a \textit{minimal meaningful line segment} under the Helmholtz principle and the \textit{a contrario} model \cite{meaningful00}:
\begin{align}
l_{\min}
=
\frac{\log(\varepsilon)-\log(N_L)}
{\log(p)},
\end{align}
where $\varepsilon$ is the expected number of tolerated false detections, 
$N_L$ is the number of possible line-segment hypotheses in the image, 
and $p$ is the probability that a random pixel is orientation-compatible with a line hypothesis.
For an image of size $M\times N$, 
we set $\varepsilon=1$,
$N_L=\sqrt{(M\cdot N)^5}$, 
and $p=3/16$.
The latter follows from using 16 uniformly sampled orientations together with $\tau=\pi/16$.
Further details of the meaningful-alignment formulation are given in \cite{meaningful00}.

Instead of counting all supporting pixels equally, 
ADLA defines the effective support $l_{\mathrm{set}}$ as
\begin{align}
l_{\mathrm{set}}
=
\sum_{i=1}^{N_s} p_w(x(i),y(i)),
\end{align}
where
\begin{align}
p_w(x(i),y(i))
=
\begin{cases}
1.0, & \text{if } p(x(i),y(i))\geq0.3,\\
p(x(i),y(i)), & \text{otherwise.}
\end{cases}
\end{align}
Thus, sufficiently strong edge responses contribute fully to the support measure, 
whereas weaker responses contribute proportionally to their ESM values.
This design allows weak but geometrically consistent evidence to reinforce a candidate while limiting its influence on validation.

For a support set satisfying $l_{\mathrm{set}}\geq l_{\min}$, 
the final segment is estimated from the refined weighted centroid and line orientation.
The two endpoints are obtained by projecting the supporting pixels onto the estimated line and selecting the largest extents on the two sides of the centroid.
The resulting line segment therefore summarizes both the spatial distribution and the confidence of the accumulated edge evidence.

Overall, 
ADLA uses edge strength information at several stages---
orientation estimation, seed selection, processing order, adaptive model refinement, and segment validation---
while its geometric aggregation itself remains independent of a particular annotation paradigm.
This combination is what allows the framework to use task-appropriate edge evidence while retaining a common line-extraction mechanism across generic, wireframe, and Manhattan line segment detection.

%% file: experiments_revised.tex
\section{Experiments}
\label{experiments}
We evaluate ADLA under three annotation paradigms: 
generic line segments, wireframe line segments, and Manhattan-world line segments.  
This design tests whether the same geometric aggregation mechanism remains effective 
when the definition of a valid line segment changes.

\subsection{Datasets, baselines, and implementation details}
\label{experimental_setup}

The YorkUrban-LineSegment dataset \cite{linelet} provides exhaustive annotations of perceptually meaningful line segments for all 102 images in the YorkUrban dataset. 
In contrast, 
the original YorkUrban annotations \cite{3D_line} retain only line segments consistent with the Manhattan-world assumption. 
Because these two benchmarks share the same indoor and outdoor images 
but adopt different annotation protocols, 
they enable a controlled comparison between generic and Manhattan line-segment detection. 
The ShanghaiTech wireframe dataset \cite{dwp} contains 5,462 images of man-made environments, 
comprising 5,000 training images and 462 test images.

We compare ADLA with the generic line-segment detectors LSD \cite{lsd}, Linelet \cite{linelet}, DeepLSD \cite{deeplsd}, and AAGLSD \cite{aaglsd}, 
as well as the wireframe detectors HAWPv2, HAWPv3 \cite{hawpv2}, ScaleLSD \cite{scalelsd}, DT-LSD \cite{dtlsd}, and LINEA \cite{linea}. 
We use the default configurations for LSD, Linelet, and AAGLSD 
and the publicly released checkpoints for all learned baselines. 
For methods requiring output filtering, 
HAWPv2, HAWPv3, and LINEA retain line segments whose predicted confidence exceeds $\lambda_c$, 
whereas ScaleLSD retains line segments supported by more than $\lambda_s$ pixels. 
We report representative operating points obtained by varying these thresholds, 
thereby exposing the corresponding precision--recall trade-offs 
instead of characterizing each method using a single, potentially favorable threshold.

All running times are measured on a workstation equipped with a 13th Gen Intel Core i9-13900K CPU, 
64~GB of memory, 
and an NVIDIA RTX~4090D GPU. 
On the 462 ShanghaiTech test images, 
whose average resolution is approximately $500\times390$ pixels, 
CPU-based ADLA processes the precomputed edge-strength maps (ESMs) in 14.6~s. 
This corresponds to 31.6~ms per image, or 31.6 images/s. 
On the same test set, 
GPU-based EdgeNAT requires 41.24~s for single-scale inference 
and approximately 535~s for multi-scale inference. 
Consequently, 
the complete sequential EdgeNAT+ADLA pipeline requires 55.84~s with single-scale EdgeNAT, 
corresponding to 120.9~ms per image or 8.27 images/s. 
With multi-scale EdgeNAT, 
the complete pipeline requires approximately 549.6~s, 
corresponding to 1.19~s per image or 0.84 images/s.

\subsection{Evaluation protocol}
\label{evaluation_metric}
We use the heatmap-based protocol adopted by prior line-segment detectors 
because it evaluates the spatial agreement between detected and annotated line support at the original image resolution. 
We do not report structural average precision (sAP): 
sAP evaluates endpoint-level segment matching and ranking, 
whereas the present study focuses on the localization and coverage of rasterized line evidence across detectors 
with different output representations and confidence mechanisms.

For each image $I$, 
the detected and ground-truth segments are independently rasterized into binary maps $D_I$ and $G_I$.  
A segment of Euclidean length $L$ is sampled at $\lceil L\rceil$ evenly spaced positions, 
including its endpoints;
the sample coordinates are rounded to the nearest image pixel 
and clipped to the image boundary.  
Correspondences between $D_I$ and $G_I$ are then computed using the standard one-to-one pixel-matching procedure with a maximum matching distance of $0.01$ times the image diagonal.  
This tolerance accommodates rasterization and small localization errors 
while preventing a pixel from being counted repeatedly.

Let $M_I^D$ and $M_I^G$ denote the matched detected and ground-truth pixels, respectively.  
Image-level precision and recall are
\begin{align}
P_I &= \frac{|M_I^D|}{|D_I|+\epsilon}, &
R_I &= \frac{|M_I^G|}{|G_I|+\epsilon},
\end{align}
where $\epsilon=10^{-15}$ prevents division by zero.  
The heatmap precision and recall are the unweighted averages over the $N_{test}$ test images,
\begin{align}
\mathrm{AP}^{\mathrm H} &= \frac{1}{N_{test}}\sum_{i=1}^{N_{test}}P_i, &
\mathrm{AR}^{\mathrm H} &= \frac{1}{N_{test}}\sum_{i=1}^{N_{test}}R_i,
\end{align}
and their harmonic mean is
\begin{align}
\mathrm{F}^{\mathrm H} =
\frac{2\,\mathrm{AP}^{\mathrm H}\mathrm{AR}^{\mathrm H}}
{\mathrm{AP}^{\mathrm H}+\mathrm{AR}^{\mathrm H}}.
\end{align}
Here, $\mathrm{AP}^{\mathrm H}$ denotes average \emph{per-image pixel precision}. 
The same protocol and original image resolution are used for all methods.

\subsection{Parameter analysis}
ADLA uses the orientation-window radius $W$, 
the number of discrete orientations $P$, 
the angular tolerance $\tau$, 
the orthogonal-distance threshold $l_w$, 
the seed threshold $\lambda$, 
and the search-neighborhood size $s\times s$.  
Unless stated otherwise, 
we set $W=7$, 
$P=16$, 
$\tau=\pi/16$, 
and $l_w=3$, 
and analyze the two task-dependent parameters $\lambda$ and $s$ below.

\subsubsection{Influence of $\lambda$}
The threshold $\lambda$ controls which ESM pixels may initialize aggregation.
We vary $\lambda$ from 0 to 0.9 in increments of 0.1, 
using $s=5$ on YorkUrban-LineSegment dataset 
and $s=3$ on ShanghaiTech dataset.  
Table~\ref{lambda} shows that
$\mathrm{F}^{\mathrm H}$ is stable over a broad range on both datasets.  
Lower thresholds are preferable under generic annotations, 
where weak but meaningful segments should be retained.  
Under wireframe annotations, 
the best score is obtained at a higher threshold 
because suppressing non-structural edge evidence improves precision.

\begin{table}[htbp]
  \centering
  \caption{Influence of $\lambda$ on the performance of ADLA on the YorkUrban-LineSegment dataset (Generic) and ShanghaiTech dataset (Wireframe).}
  \label{lambda}
  \renewcommand{\arraystretch}{0.8}   
  \setlength{\tabcolsep}{3pt}
  \begin{tabular}{c c c c c c c c c c c}
    \toprule
    Dataset ($\lambda$) & 0.0 & 0.1 & 0.2 & 0.3 & 0.4 & 0.5 & 0.6 & 0.7 & 0.8 & 0.9 \\
    \midrule
    Generic ($\mathrm{F}^{\mathrm{H}}$) & 0.8467 & 0.8468 & 0.8466 & 0.8467 & 0.8464 & 0.8450 & 0.8426 & 0.8381 & 0.8294 & 0.7729 \\
    \midrule
Wireframe ($\mathrm{F}^{\mathrm{H}}$) & 0.8590 & 0.8590 & 0.8591 & 0.8597 & 0.8632 & 0.8669 & 0.8697 & 0.8716 & 0.8720 & 0.8674 \\
    \bottomrule
    \end{tabular}
  \end{table}

\subsubsection{Influence of $s$}
The neighborhood size determines how far ADLA searches around the current support set for additional pixels.  
We vary $s$ with $\lambda=0.1$ on YorkUrban-LineSegment dataset 
and $\lambda=0.8$ on ShanghaiTech dataset.  
As shown in Table~\ref{s_york}, 
increasing $s$ improves generic-line recall and hence $\mathrm{F}^{\mathrm H}$, 
although the accompanying precision decrease is examined in the full comparison below.  
On ShanghaiTech dataset, 
the score changes by at most 0.0012 for $s\in\{3,5,7\}$, 
indicating little sensitivity in this range.

\begin{table}[htbp]
  \centering
  \caption{Influence of $s \times s$ on the performance of ADLA on the YorkUrban-LineSegment dataset (Generic) and ShanghaiTech dataset (Wireframe).}
  \label{s_york}
  \renewcommand{\arraystretch}{0.8}   
  \setlength{\tabcolsep}{3pt}
  \begin{tabular}{c c c c c c}
    \toprule
   Dataset ($s\times s$) & $3\times 3$ & $5\times 5$ & $7\times 7$ & $11\times 11$ & $15 \times 15$ \\
    \midrule
    Generic ($\mathrm{F}^{\mathrm{H}}$) & 0.8209 & 0.8468 & 0.8535 & 0.8619 & 0.8665 \\
    \midrule
    Wireframe ($\mathrm{F}^{\mathrm{H}}$) & 0.8720 & 0.8722 & 0.8712 & - & - \\
    \bottomrule
    \end{tabular}
  \end{table}

\subsection{Ablation study}
\label{ablation_study}
We ablate the two geometric constraints and the adaptive line-model refinement
to isolate their respective contributions.  
Three progressively more complete variants are evaluated: 
\textsc{Ori}, 
which aggregates pixels using orientation coherence alone; 
\textsc{Ori+Dist}, 
which additionally enforces the bounded orthogonal-distance constraint 
using the line model initialized by the seed; 
and the complete ADLA, 
which further updates the edge-strength-weighted centroid and orientation during aggregation.  
All remaining components, including ESMs, seed selection, processing order, meaningful-support validation, and endpoint estimation, 
are kept unchanged.  
We use the same operating settings as the corresponding $s=5$ experiments: 
$\lambda=0.1$ on YorkUrban-LineSegment dataset 
and $\lambda=0.8$ on ShanghaiTech dataset.

In addition to $\mathrm{AP}^{\mathrm H}$, $\mathrm{AR}^{\mathrm H}$, and $\mathrm{F}^{\mathrm H}$, 
we report the average number of detected segments per image ($\overline{N}_{\mathrm{seg}}$) 
and their average length in pixels ($\overline{L}_{\mathrm{seg}}$).  
The latter two statistics are not accuracy metrics; 
they diagnose fragmentation, 
which can be obscured by a rasterized heatmap score 
because several short collinear detections may cover much of the same support as one longer segment.

\begin{table}[htbp]
  \centering
  \caption{Ablation study on YorkUrban-LineSegment dataset (generic) and ShanghaiTech dataset (wireframe).  $\overline{N}_{\mathrm{seg}}$ is the average number of detected segments per image, and $\overline{L}_{\mathrm{seg}}$ is their average length in pixels.}
  \label{tab:ablation}
  \small
  \renewcommand{\arraystretch}{0.8}
  \setlength{\tabcolsep}{3pt}
  \begin{tabular}{l l c c c c c}
    \toprule
    Dataset & Variant & $\mathrm{AP}^{\mathrm H}$ & $\mathrm{AR}^{\mathrm H}$ & $\mathrm{F}^{\mathrm H}$ & $\overline{N}_{\mathrm{seg}}$ & $\overline{L}_{\mathrm{seg}}$ \\
    \midrule
    Generic & \textsc{Ori}      & 0.9213 & 0.7515 & 0.8278 & 326 & 58.93 \\
    Generic & \textsc{Ori+Dist} & 0.9265 & 0.7726 & 0.8426 & 431 & 44.04 \\
    Generic & ADLA & \textbf{0.9277} & \textbf{0.7788} & \textbf{0.8468} & 380 & 50.92 \\
    \midrule
    Wireframe & \textsc{Ori}      & 0.8364 & 0.8637 & 0.8498 & 66  & 87.65 \\
    Wireframe & \textsc{Ori+Dist} & 0.8480 & 0.8804 & 0.8639 & 104 & 54.39 \\
    Wireframe & ADLA              & \textbf{0.8533} & \textbf{0.8921} & \textbf{0.8722} & 85 & 69.64 \\
    \bottomrule
  \end{tabular}
\end{table}

Table~\ref{tab:ablation} shows consistent gains as the proposed components are introduced.  
Adding the bounded-distance constraint to orientation coherence
raises $\mathrm{F}^{\mathrm H}$ from 0.8278 to 0.8426 on YorkUrban-LineSegment dataset, 
and from 0.8498 to 0.8639 on ShanghaiTech dataset.  
Thus, local orientation agreement alone is insufficient: 
explicitly constraining support to a narrow band around the line hypothesis 
improves both precision and recall.
However, 
\textsc{Ori+Dist} also produces the most segments and the shortest average length on both datasets.  
This indicates that a fixed seed-initialized model tends to split extended support 
when its initial geometry does not remain representative of the growing segment.

Adaptive refinement resolves much of this fragmentation.  
Relative to \textsc{Ori+Dist}, 
complete ADLA reduces the average number of segments from 431 to 380 and increases their average length from 44.04 to 50.92 pixels on YorkUrban-LineSegment dataset; 
on ShanghaiTech dataset, 
the corresponding changes are from 104 to 85 segments and from 54.39 to 69.64 pixels.  
At the same time, 
it further improves $\mathrm{F}^{\mathrm H}$ to 0.8468 and 0.8722, respectively.  
These results support the central design of ADLA: 
the two constraints improve the geometric reliability of aggregation, 
while updating the line model from the accumulated edge evidence permits longer, less fragmented segments without sacrificing detection accuracy.

\begin{table}[htbp]
  \centering
  \caption{Quantitative comparisons of different algorithms on the YorkUrban-LineSegment dataset.}
  \label{york_compare}
  \renewcommand{\arraystretch}{0.8}   
  \setlength{\tabcolsep}{8pt}
  \begin{tabular}{c c c c}
    \toprule
    Methods & $\mathrm{AP}^{\mathrm{H}}$ & $\mathrm{AR}^{\mathrm{H}}$ & $\mathrm{F}^{\mathrm{H}}$ \\
    \midrule
    LSD \cite{lsd} & 0.9077 & 0.7354 & 0.8125 \\
    \midrule
    Linelet \cite{linelet} & 0.8595 & 0.8428 & 0.8511 \\
    \midrule
   DeepLSD \cite{deeplsd} & 0.8818 & 0.7708 & 0.8226 \\
   \midrule
   HAWPv2 ($\lambda_c=0.4$) \cite{hawpv2} & 0.8731 & 0.6612 & 0.7525\\
   \midrule
   HAWPv2 ($\lambda_c=0.5$) \cite{hawpv2} & 0.9075 & 0.6093 & 0.7291 \\
   \midrule
   HAWPv2 ($\lambda_c=0.6$) \cite{hawpv2} & 0.9346 & 0.5566 & 0.6977\\
   \midrule
   HAWPv2 ($\lambda_c=0.8$) \cite{hawpv2} & 0.9720 & 0.4124 & 0.5791\\
   \midrule
   HAWPv3 ($\lambda_c=0.1$) \cite{hawpv2} & 0.9092 & 0.4083 & 0.5636\\
\midrule
HAWPv3 ($\lambda_c=0.2$) \cite{hawpv2} & 0.9373 & 0.3397 & 0.4987\\
\midrule
HAWPv3 ($\lambda_c=0.3$) \cite{hawpv2} & 0.9519 & 0.2896 & 0.444\\
\midrule
ScaleLSD ($\lambda_s=5$) \cite{scalelsd} & 0.8946 & 0.5389 & 0.6726 \\
\midrule
ScaleLSD ($\lambda_s=10$) \cite{scalelsd} & 0.9323 & 0.4572 & 0.6135 \\
\midrule
ScaleLSD ($\lambda_s=15$) \cite{scalelsd} & 0.9466 & 0.4156 & 0.5776 \\
\midrule
ScaleLSD ($\lambda_s=20$) \cite{scalelsd} & 0.9558 & 0.3826 & 0.5465 \\
\midrule
ScaleLSD ($\lambda_s=30$) \cite{scalelsd} & 0.9638 & 0.3232 & 0.4840 \\
   \midrule
   DT-LSD \cite{dtlsd} & 0.9563 & 0.5866 & 0.7272 \\
   \midrule
   LINEA ($\lambda_c=0.15$) \cite{linea} & 0.8089 & 0.8015 & 0.8052 \\
\midrule
LINEA ($\lambda_c=0.20$) \cite{linea} & 0.9143 & 0.6633 & 0.7688 \\
\midrule
LINEA ($\lambda_c=0.25$) \cite{linea} & 0.9591 & 0.5329 & 0.6851 \\
   \midrule
   AAGLSD \cite{aaglsd} & 0.9334 & 0.6037 & 0.7332 \\
    \midrule
    ADLA$_{5\times5}$ & 0.9277 & 0.7788 & 0.8468 \\
    \midrule
    ADLA$_{7\times 7}$ & 0.9209 & 0.7952 & 0.8535 \\
    \midrule
    ADLA$_{15\times 15}$ & 0.8960 & 0.8389 & 0.8665 \\
    \bottomrule
    \end{tabular}
  \end{table}

\subsection{Generic line segment detection on YorkUrban-LineSegment dataset}
We set $\lambda=0.1$ and report ADLA with $s\in\{5,7,15\}$, 
denoted by ADLA$_{5\times5}$, ADLA$_{7\times7}$, and ADLA$_{15\times15}$, respectively.
Table~\ref{york_compare} compares these configurations with the generic and wireframe baselines.

\begin{figure}[pos=htbp]
  \centering
  \begin{tabular}{ccc}
    \includegraphics[width=0.3\textwidth]{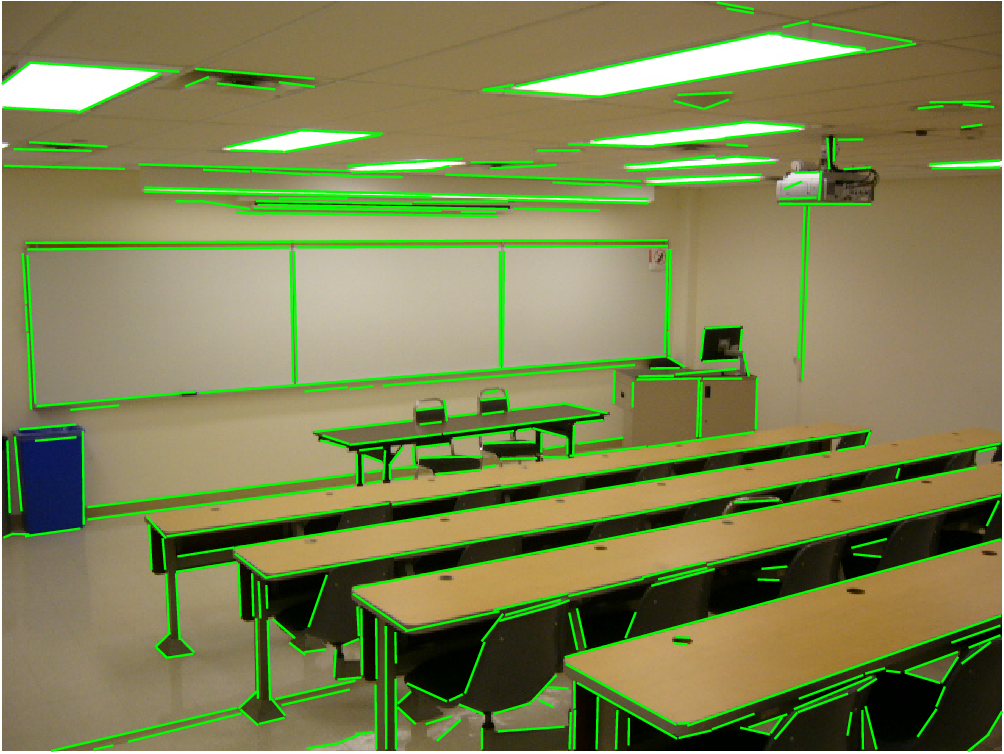} &
    \includegraphics[width=0.3\textwidth]{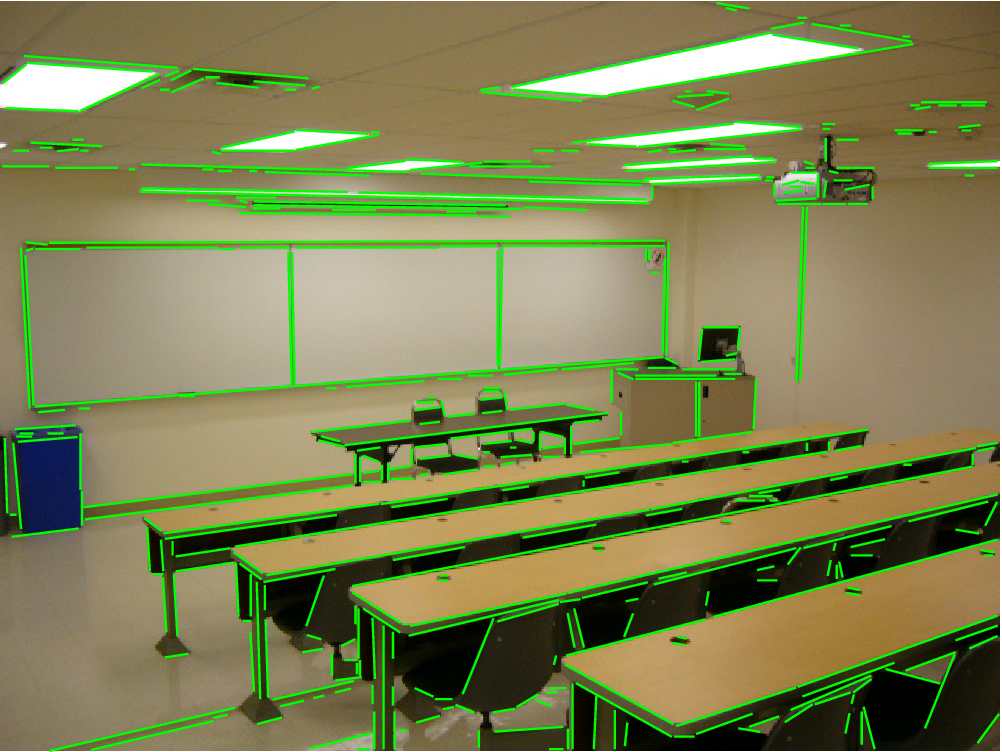} &
    \includegraphics[width=0.3\textwidth]{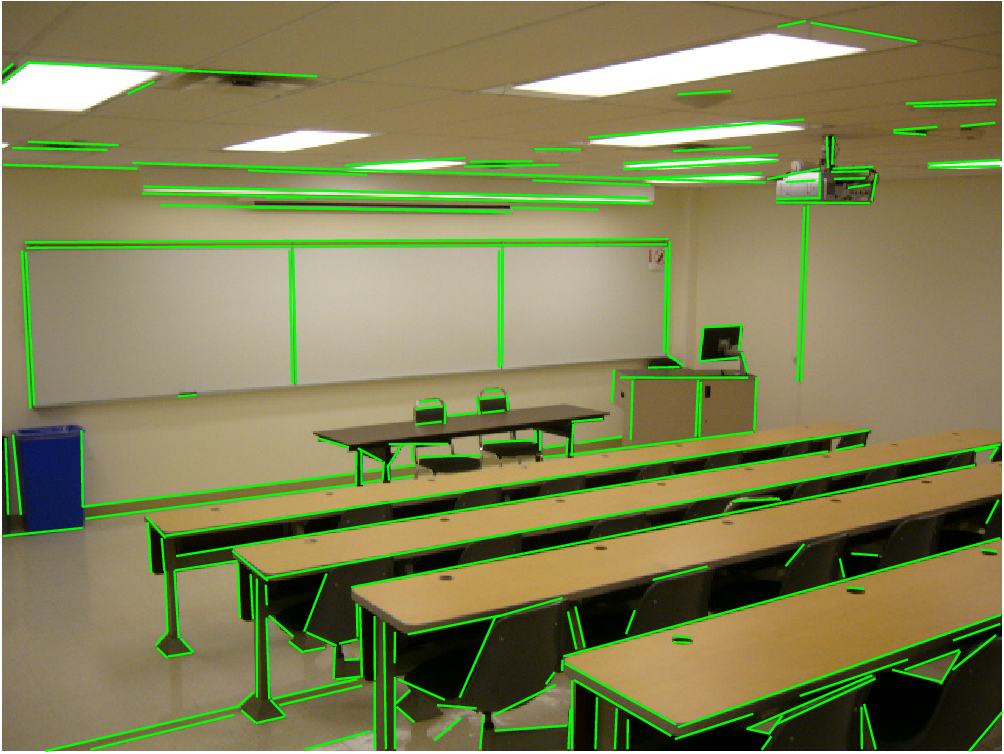} \\
    (a) LSD & (b) Linelet & (c) DeepLSD \\
        \includegraphics[width=0.3\textwidth]{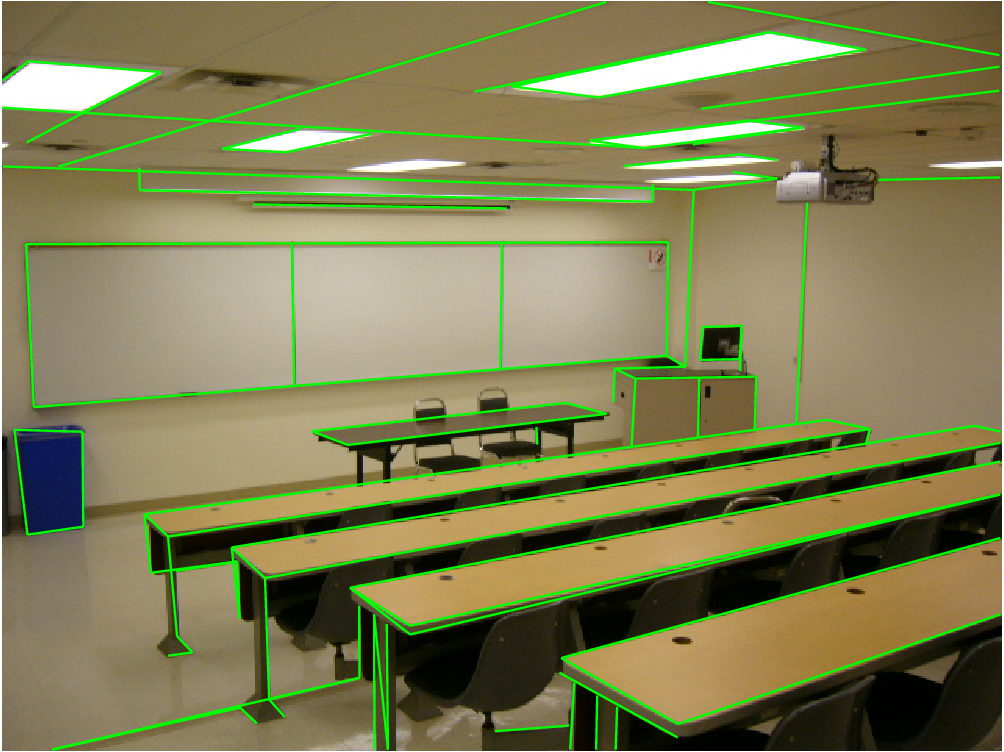} &
    \includegraphics[width=0.3\textwidth]{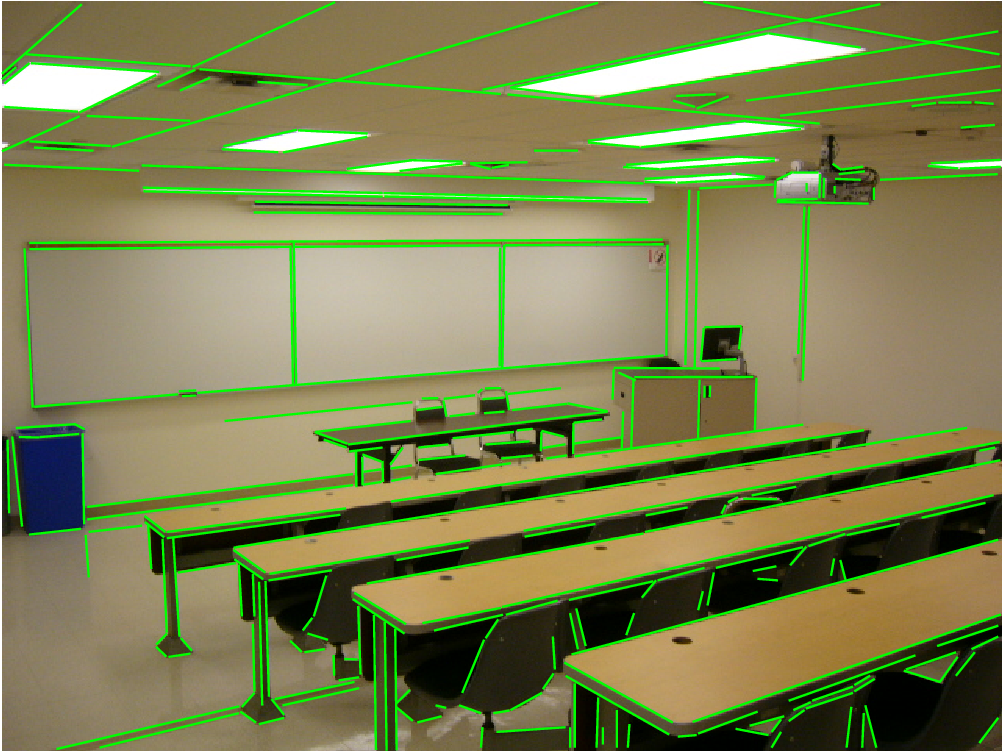} &
        \includegraphics[width=0.3\textwidth]{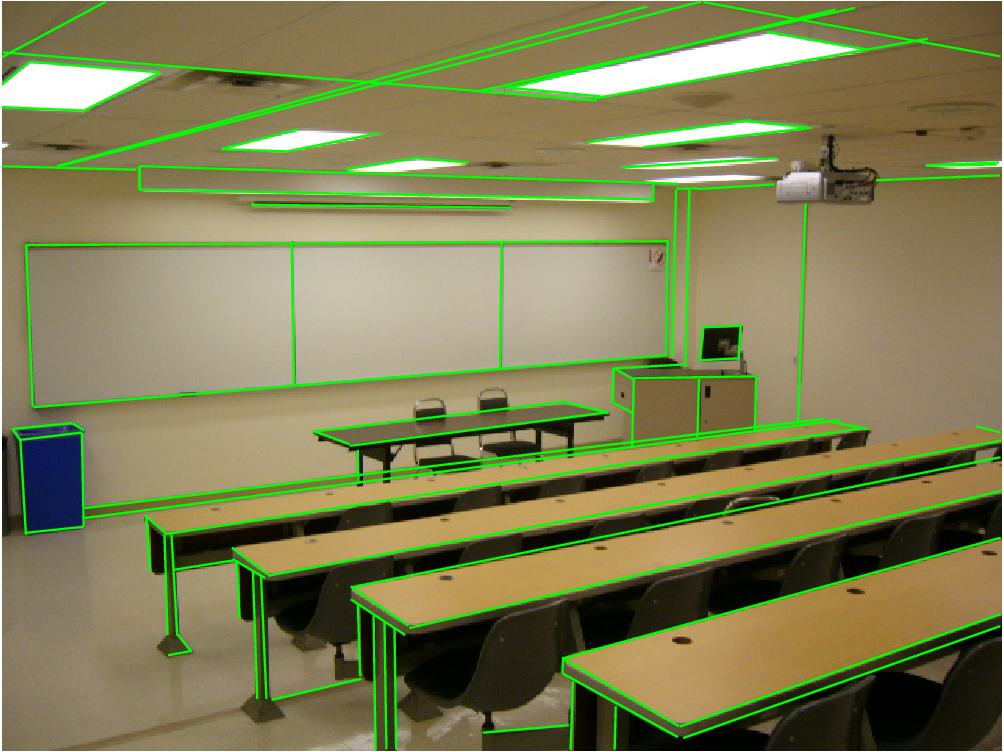} \\
(d) HAWPv2 & (e) Ground truth & (f) DT-LSD  \\
    \includegraphics[width=0.3\textwidth]{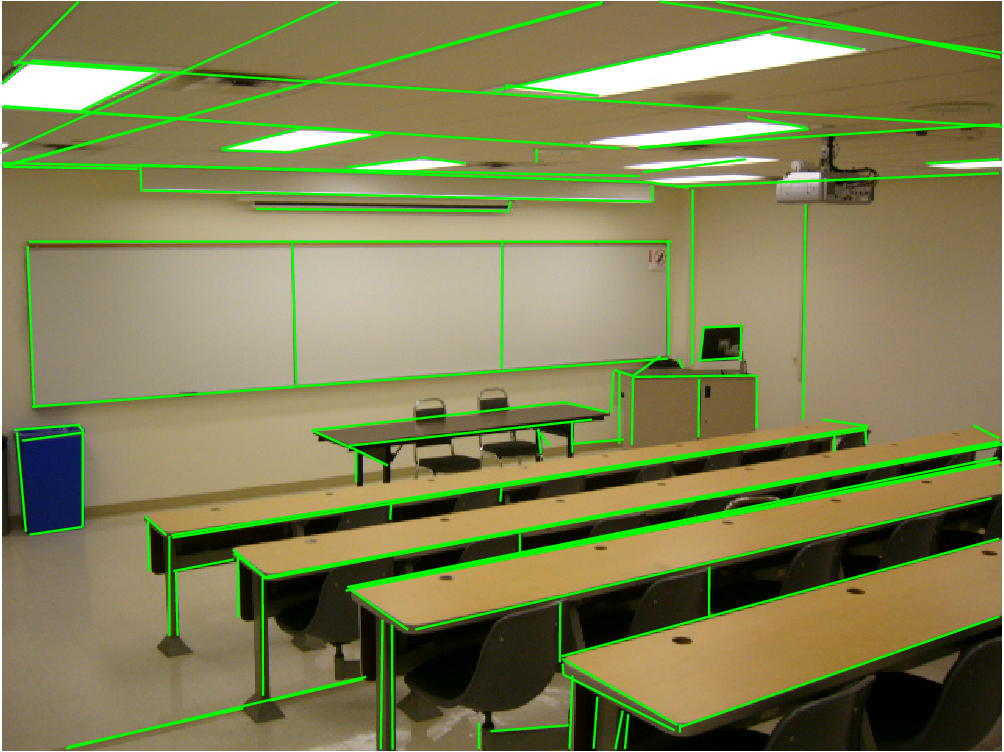} &
    \includegraphics[width=0.3\textwidth]{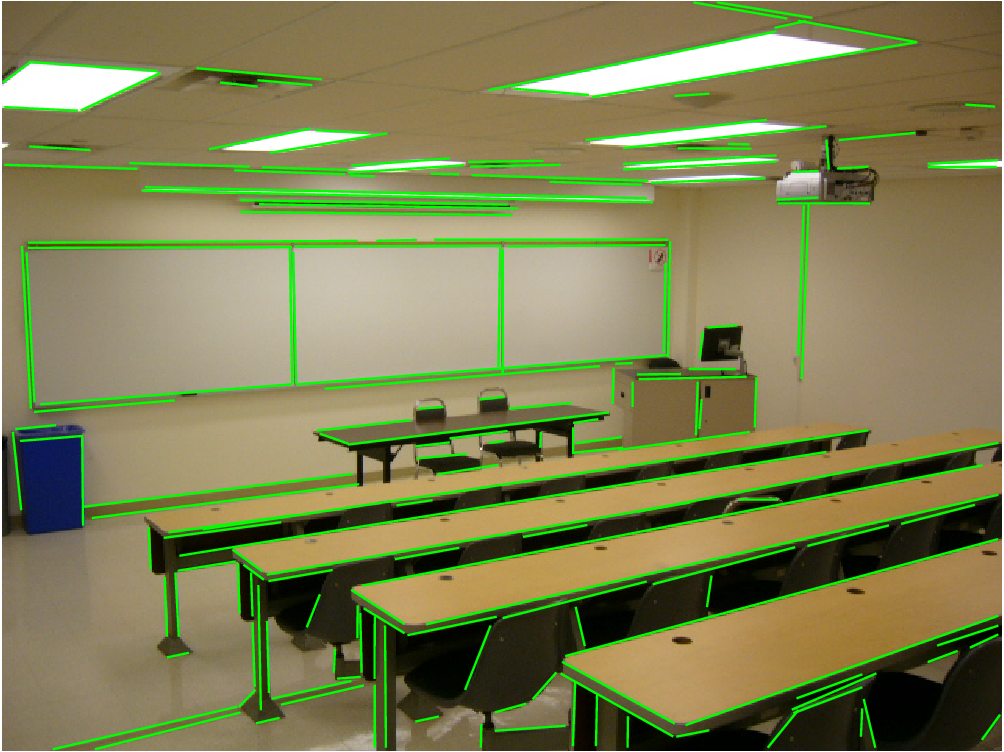} &
    \includegraphics[width=0.3\textwidth]{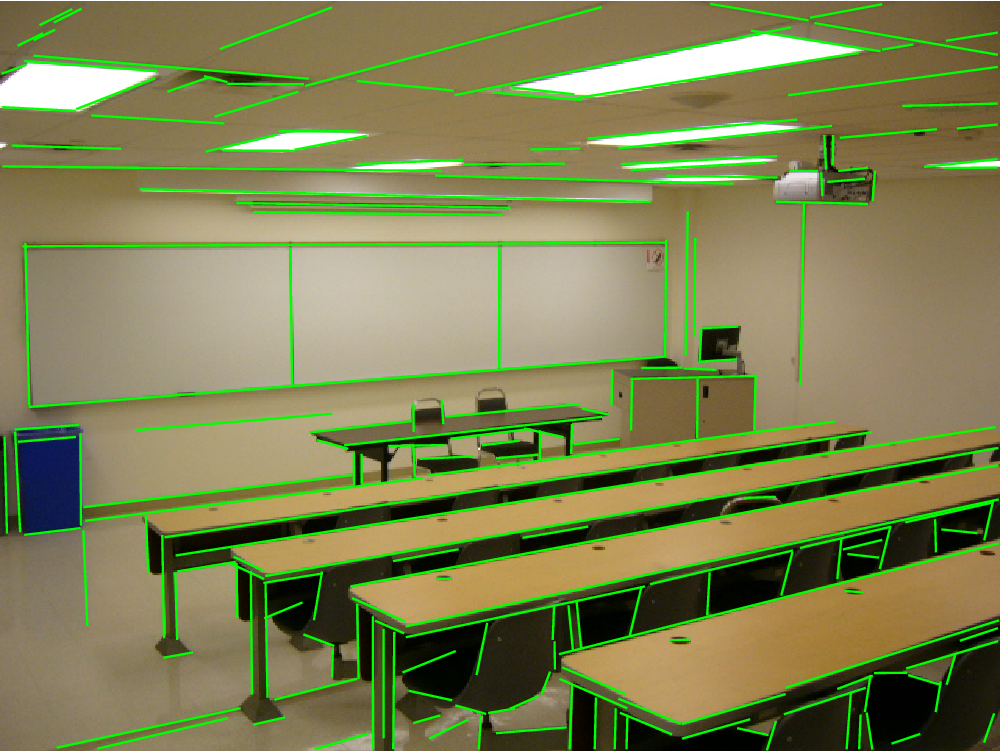} \\
(g) LINEA & (h) AAGLSD & (i) ADLA$_{5 \times 5}$ \\
  \end{tabular}
  \caption{Line segment detection results computed by different methods on an image from the YorkUrban-LineSegment dataset}
  \label{york_image}
  \end{figure}

ADLA$_{15\times15}$ obtains the highest $\mathrm{F}^{\mathrm H}$, 0.8665,
exceeding the strongest baseline, Linelet (0.8511), by 1.54 percentage points.
Its larger neighborhood favors recall (0.8389) at the cost of lower precision (0.8960).  
ADLA$_{7\times7}$ provides a more balanced operating point: 
it improves over Linelet in both precision (0.9209 versus 0.8595) and $\mathrm{F}^{\mathrm H}$ (0.8535 versus 0.8511).  
ADLA$_{5\times5}$ yields still higher precision (0.9277) with a small reduction in recall.  
The wireframe detectors can attain high precision on this benchmark, 
but their substantially lower recall confirms that 
models optimized for structural subsets omit many segments included by the generic annotations.

Figure \ref{york_image} illustrates representative results. 
LSD and Linelet are more fragmented; 
LSD also misses segments, 
whereas Linelet introduces more false detections. 
DeepLSD reduces fragmentation but misses true segments 
and sometimes produces overlapping detections. 
The wireframe-trained HAWPv2, DT-LSD, and LINEA omit many generically annotated segments, 
as does AAGLSD. 
ADLA recovers more true segments while limiting false detections, 
although its output is less structurally complete than that of specialized wireframe detectors.

\subsection{Wireframe line segment detection on ShanghaiTech dataset}
We set $s=3$ and report ADLA at $\lambda\in\{0.8,0.9\}$ to show two precision--recall operating points.  
Table~\ref{wireframe_compare} includes both scores produced with our evaluation code 
and previously published scores where available.  
An asterisk marks values taken from HAWPv2 \cite{hawpv2}, 
except for DT-LSD \cite{dtlsd} and EM-LSD \cite{emlsd}, 
whose marked values are taken from their respective papers.

\begin{table}[htbp]
  \centering
  \caption{Quantitative comparisons of different algorithms on the ShanghaiTech dataset. $^{*}$ indicates that the scores computed by the method are extracted from the paper of HAWPv2, DT-LSD, and EM-LSD.}
  \label{wireframe_compare}
  \renewcommand{\arraystretch}{0.8}   
  \setlength{\tabcolsep}{8pt}
  \begin{tabular}{c c c c}
    \toprule
    Methods & $\mathrm{AP}^{\mathrm{H}}$ & $\mathrm{AR}^{\mathrm{H}}$ & $\mathrm{F}^{\mathrm{H}}$ \\
    \midrule
    LSD \cite{lsd} & 0.4993 & 0.8502 & 0.6291 \\
    \midrule
    Linelet \cite{linelet} & 0.4499 & 0.89 & 0.5977 \\
    \midrule
 DeepLSD \cite{deeplsd} & 0.4817 & 0.8527 & 0.6156 \\
 \midrule
 DWP$^{*}$ \cite{dwp} & 0.678 & - & 0.722 \\
 \midrule
 AFM$^{*}$ \cite{afm} & 0.692 & - & 0.772 \\
 \midrule
 AFM++$^{*}$ \cite{afmjournal} & 0.748 & - & 0.828\\
 \midrule
 L-CNN$^{*}$ \cite{lcnn} & 0.816 & - & 0.779 \\
 \midrule
 F-Clip$^{*}$ (HG2-LB) \cite{fclip} & 0.851 & - & 0.809\\
 \midrule
 LETR$^{*}$ (R101) \cite{letr} & 0.855 & - & 0.798 \\
 \midrule
 LETR$^{*}$ (R50) \cite{letr} & 0.847 & - & 0.791 \\
 \midrule
 ELSD$^{*}$ (HG) \cite{elsd} & 0.847 & - & 0.803\\
 \midrule
 ELSD$^{*}$ (Res34) \cite{elsd} & 0.872 & - & 0.823 \\
 \midrule
 HAWPv1$^{*}$ \cite{hawp} & 0.845 & - & 0.803 \\
 \midrule
 HAWPv2$^{*}$ \cite{hawpv2} & 0.880 & - & 0.814 \\
 \midrule
   HAWPv2 ($\lambda_c=0.7$) \cite{hawpv2} & 0.7866 & 0.8496 & 0.8169\\
   \midrule
   HAWPv2 ($\lambda_c=0.8$) \cite{hawpv2} & 0.8428 & 0.7917 & 0.8165\\
   \midrule
   HAWPv2 ($\lambda_c=0.9$) \cite{hawpv2} & 0.9069 & 0.6701 & 0.7707 \\
   \midrule
 HAWPv3 ($\lambda_c=0.7$) \cite{hawpv2} & 0.7832 & 0.3236 & 0.4580\\
 \midrule
 HAWPv3 ($\lambda_c=0.8$) \cite{hawpv2} & 0.7984 & 0.2473 & 0.3777\\
 \midrule
 HAWPv3 ($\lambda_c=0.9$) \cite{hawpv2} & 0.8214 & 0.1563 & 0.2626\\
 \midrule
 ScaleLSD ($\lambda_s=20$) \cite{scalelsd} & 0.6078 & 0.6518 & 0.6290 \\
 \midrule
 ScaleLSD ($\lambda_s=40$) \cite{scalelsd} & 0.6735 & 0.5051 & 0.5772 \\
 \midrule
 ScaleLSD ($\lambda_s=70$) \cite{scalelsd} & 0.7407 & 0.3195 & 0.4464 \\
 \midrule
 DT-LSD$^{*}$ \cite{dtlsd} & - & 0.891 & 0.858 \\
 \midrule
 DT-LSD \cite{dtlsd} & 0.8236 & 0.8813 & 0.8515 \\
 \midrule
 LINEA ($\lambda_c=0.2$) \cite{linea} & 0.6995 & 0.9249 & 0.7966 \\
 \midrule
 LINEA ($\lambda_c=0.25$) \cite{linea} & 0.8056 & 0.8630 & 0.8333 \\
 \midrule
 LINEA ($\lambda_c=0.3$) \cite{linea} & 0.8797 & 0.7758 & 0.8245 \\
 \midrule
EM-LSD$^{*}$ \cite{emlsd} & - & - & 0.802 \\
\midrule
AAGLSD \cite{aaglsd} & 0.5611 & 0.7703 & 0.6492 \\
\midrule
  ADLA ($\lambda=0.8$) & 0.8642 & 0.88 & 0.8720 \\
  \midrule
  ADLA ($\lambda=0.9$) & 0.8853 & 0.8502 & 0.8674 \\
    \bottomrule
    \end{tabular}
  \end{table}

The generic detectors obtain high recall but low precision 
because many valid generic lines are not included in the wireframe annotations.  
Among the baselines evaluated with our code, 
DT-LSD and LINEA obtain $\mathrm{F}^{\mathrm H}$ scores of 0.8515 and 0.8333, respectively.  
ADLA with $\lambda=0.8$ reaches 0.8720, 
improvements of 2.05 and 3.87 percentage points.
Increasing $\lambda$ to 0.9 raises precision from 0.8642 to 0.8853 while reducing recall from 0.8800 to 0.8502, 
illustrating the expected threshold trade-off.

\begin{figure}[pos=htbp]
\centering
\begin{tabular}{ccc}
    \includegraphics[width=0.3\textwidth]{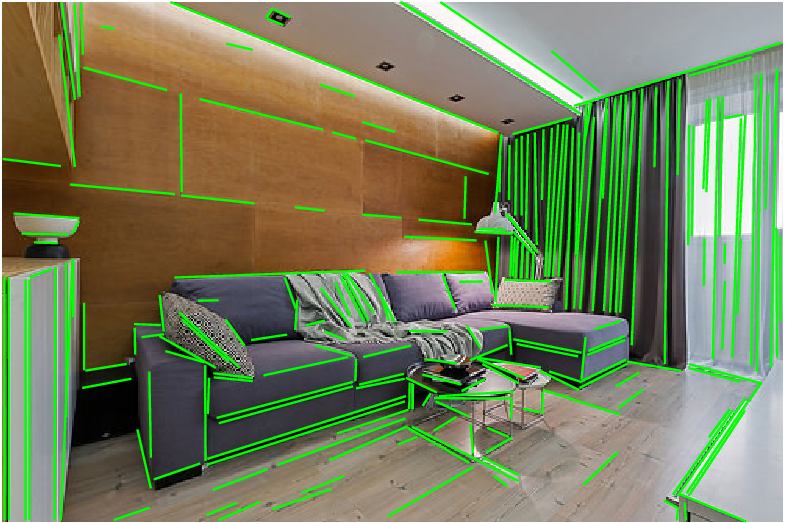} & 
    \includegraphics[width=0.3\textwidth]{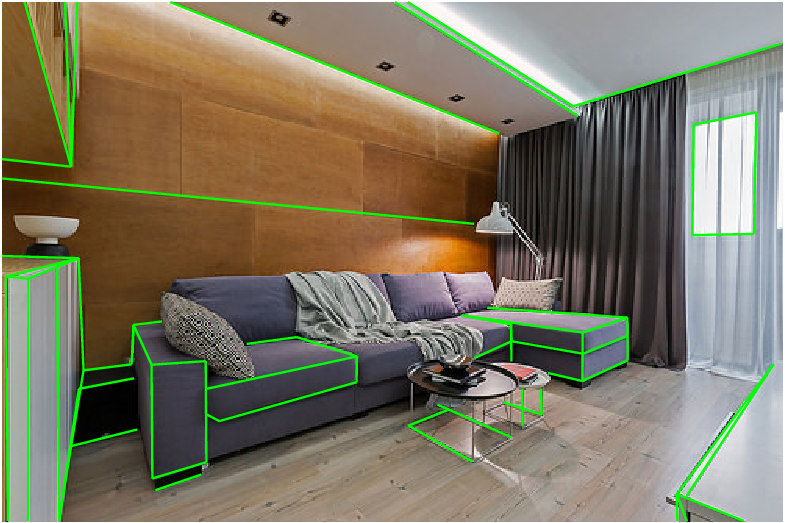} & 
    \includegraphics[width=0.3\textwidth]{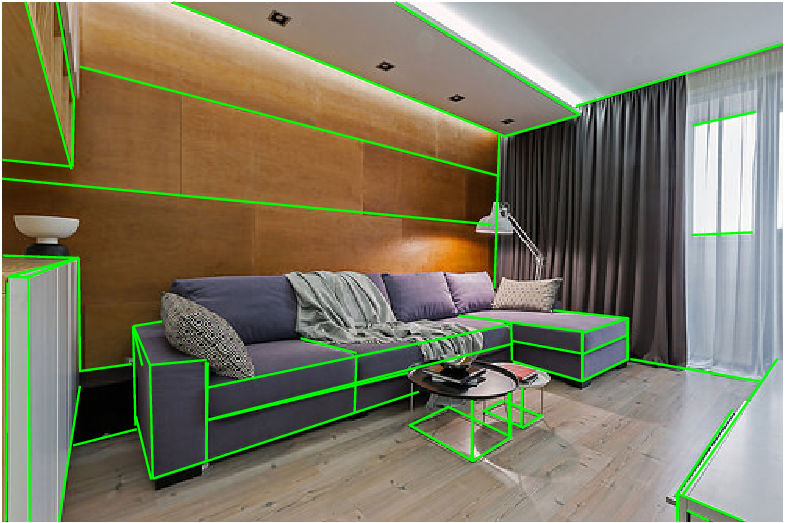} \\ 
(a) DeepLSD & (b) HAWPv2 & (c) DT-LSD \\
    \includegraphics[width=0.3\textwidth]{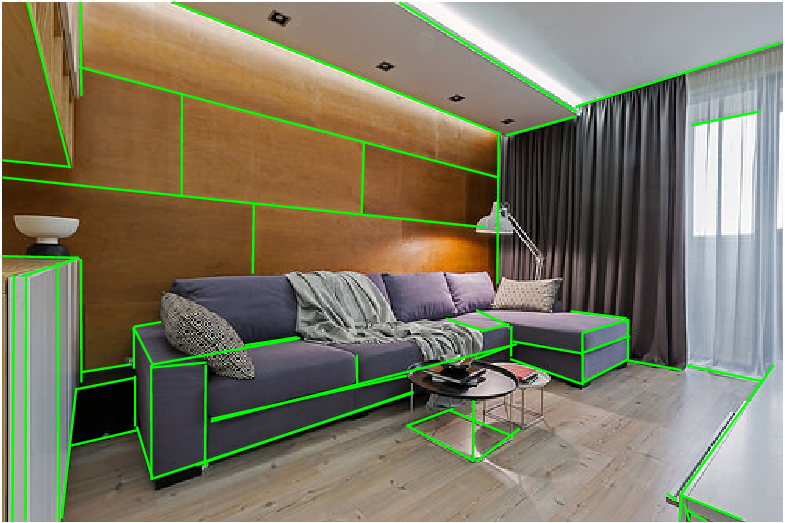} &
    \includegraphics[width=0.3\textwidth]{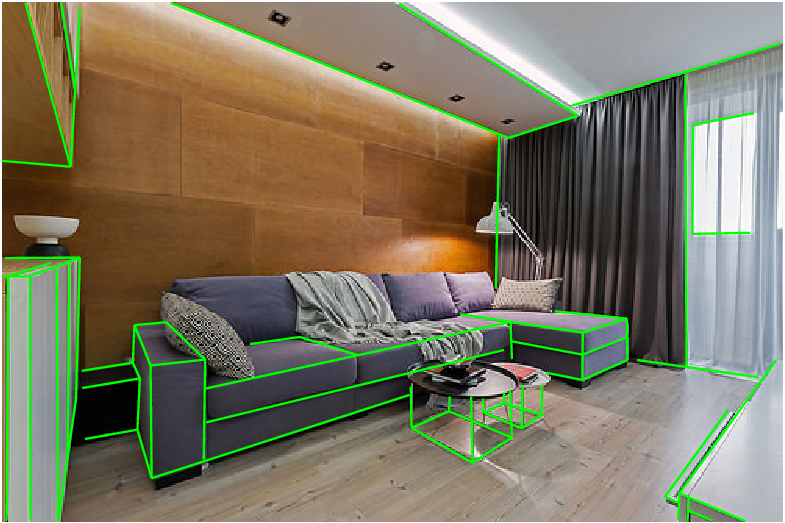} &
    \includegraphics[width=0.3\textwidth]{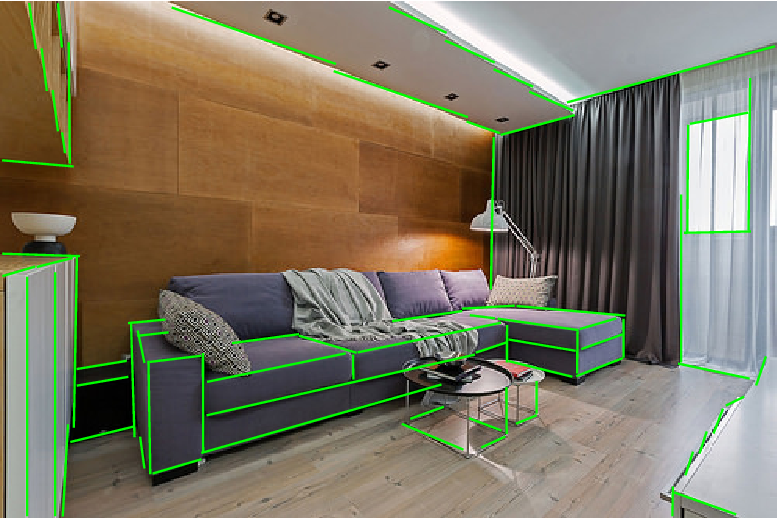} \\
(d) LINEA & (e) Ground truth  & (f) ADLA \\
\end{tabular}
  \caption{Line segment detection results computed by different methods on an image from the ShanghaiTech dataset}
  \label{wireframe_image}
\end{figure}

Figure \ref{wireframe_image} shows that DeepLSD cannot distinguish wireframe from other lines. 
HAWPv2, DT-LSD, and LINEA produce fewer false detections and more complete structures, 
although HAWPv2 and LINEA sometimes create spurious endpoint connections. 
ADLA recovers a comparable number of true segments with fewer false detections.

\subsection{Manhattan line segment detection on YorkUrban dataset}
YorkUrban dataset retains only segments satisfying the Manhattan-world assumption, 
and none of the compared methods is trained specifically for this annotation objective.  
Following existing wireframe detectors, we also provide quantitative comparisons on this dataset,
but we treat this experiment as a cross-paradigm transfer test.  
For ADLA, 
we set $\lambda=0.9$ and $s=3$ and discard ESM pixels below $\lambda$ before aggregation.  
This stricter setting suppresses lines that are valid under wireframe annotations but absent from the Manhattan ground truth.

 \begin{table}[htbp]
  \centering
  \caption{Quantitative comparisons of different algorithms on the YorkUrban dataset. $^{*}$ indicates that the scores computed by the method are extracted from the paper of HAWPv2, DT-LSD, and EM-LSD.}
  \label{york_compare_original}
  \renewcommand{\arraystretch}{0.8}   
  \setlength{\tabcolsep}{8pt}
  \begin{tabular}{c c c c}
    \toprule
    Methods & $\mathrm{AP}^{\mathrm{H}}$ & $\mathrm{AR}^{\mathrm{H}}$ & $\mathrm{F}^{\mathrm{H}}$ \\
    \midrule
LSD \cite{lsd} & 0.4333 & 0.8938 & 0.5837 \\
\midrule
Linelet \cite{linelet} & 0.3784 & 0.9397 & 0.5395 \\
\midrule
DeepLSD \cite{deeplsd} & 0.4054 & 0.9013 & 0.5592 \\
\midrule
DWP$^{*}$ \cite{dwp} & 0.510 & - & 0.616 \\
\midrule
AFM$^{*}$ \cite{afm} & 0.482 & - & 0.633 \\
\midrule
AFM++$^{*}$ \cite{afmjournal} & 0.505 & - & 0.668 \\
\midrule
L-CNN$^{*}$ \cite{lcnn} & 0.583 & - & 0.622 \\
\midrule
F-Clip$^{*}$ (HG2-LB) \cite{fclip} & 0.623 & - & 0.645 \\
\midrule
LETR$^{*}$ (R101) \cite{letr} & 0.596 & - & 0.620 \\
\midrule
LETR$^{*}$ (R50) \cite{letr} & 0.617 & - & 0.634 \\
\midrule
ELSD$^{*}$ (HG) \cite{elsd} & 0.578 & - & 0.621\\
\midrule
ELSD$^{*}$ (Res34) \cite{elsd} & 0.620 & - & 0.636 \\
\midrule
HAWPv1$^{*}$ \cite{hawp} & 0.606 & - & 0.648 \\
\midrule
HAWPv2$^{*}$ \cite{hawpv2} & 0.646 & - & 0.645\\
\midrule
HAWPv2 ($\lambda_c=0.7$) \cite{hawpv2} & 0.5822 & 0.7749 & 0.6649\\
\midrule
HAWPv2 ($\lambda_c=0.8$) \cite{hawpv2} & 0.6401 & 0.7057 & 0.6713 \\
\midrule
HAWPv2 ($\lambda_c=0.9$) \cite{hawpv2} & 0.7181 & 0.5762 & 0.6394\\
\midrule
HAWPv3 ($\lambda_c=0.4$) \cite{hawpv2} & 0.6194 & 0.4315 & 0.5086\\
\midrule
HAWPv3 ($\lambda_c=0.5$) \cite{hawpv2} & 0.6418 & 0.3768 & 0.4749\\
\midrule
HAWPv3 ($\lambda_c=0.6$) \cite{hawpv2} & 0.66 & 0.3249 & 0.4354\\
\midrule
ScaleLSD ($\lambda_s=15$) \cite{scalelsd} & 0.5642 & 0.6518 & 0.6049 \\
\midrule
ScaleLSD ($\lambda_s=40$) \cite{scalelsd} & 0.6224 & 0.4576 & 0.5274 \\
\midrule
ScaleLSD ($\lambda_s=70$) \cite{scalelsd} & 0.6543 & 0.2527 & 0.3646 \\
\midrule
DT-LSD$^{*}$ \cite{dtlsd} & - & 0.659 & 0.680 \\
\midrule
DT-LSD \cite{dtlsd} & 0.5571 & 0.8838 & 0.6834 \\
\midrule
LINEA ($\lambda_c=0.2$) \cite{linea} & 0.4850 & 0.9044 & 0.6314 \\
\midrule
LINEA ($\lambda_c=0.3$) \cite{linea} & 0.6470 & 0.7391 & 0.6900 \\
\midrule
LINEA ($\lambda_c=0.4$) \cite{linea} & 0.7338 & 0.5479 & 0.6274 \\
\midrule
EM-LSD$^{*}$ \cite{emlsd} & - & - & 0.653 \\
\midrule
AAGLSD \cite{aaglsd} & 0.4850 & 0.8073 & 0.6059 \\
\midrule
ADLA & 0.6516 & 0.8291 & 0.7297 \\
    \bottomrule
    \end{tabular}
  \end{table}

ADLA obtains an $\mathrm{F}^{\mathrm H}$ score of 0.7297, 
exceeding the best baseline evaluated with our code, LINEA at 0.6900, by 3.97 percentage points.
Its precision of 0.6516 and recall of 0.8291 indicate that 
the improvement is not obtained by simply suppressing most detections.  
Nevertheless, 
the lack of a large task-specific training set limits the conclusions that can be drawn about fully supervised Manhattan-line detection; 
the experiment primarily demonstrates the transferability of ADLA's geometric aggregation.

\begin{figure}[pos=htbp]
\centering
\begin{tabular}{ccc}
    \includegraphics[width=0.3\textwidth]{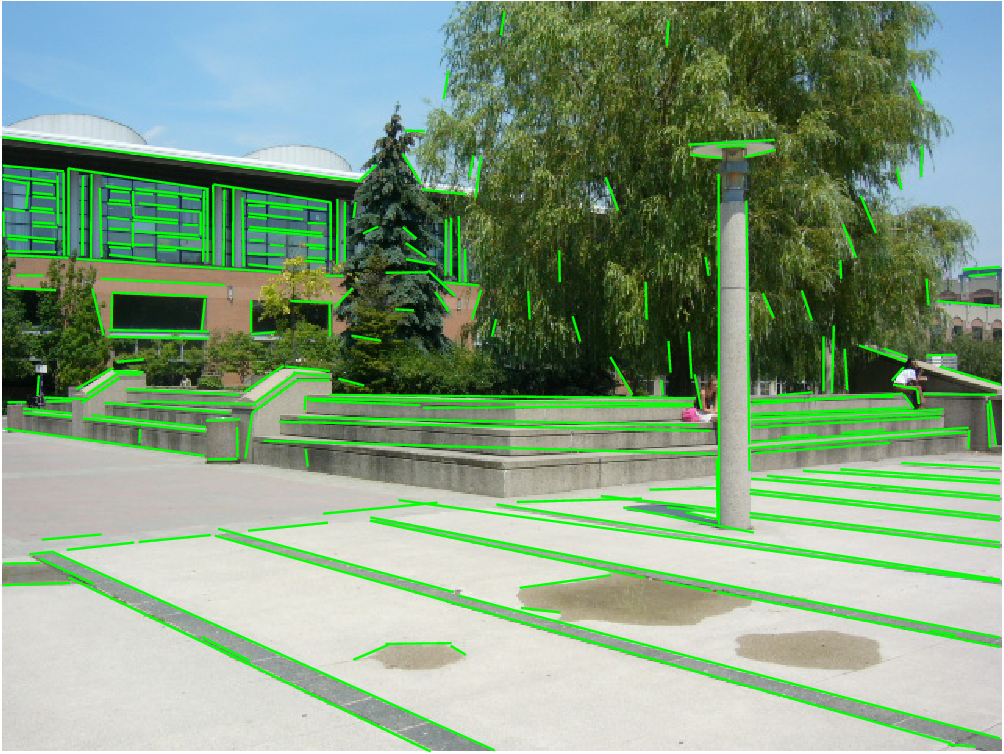} & 
    \includegraphics[width=0.3\textwidth]{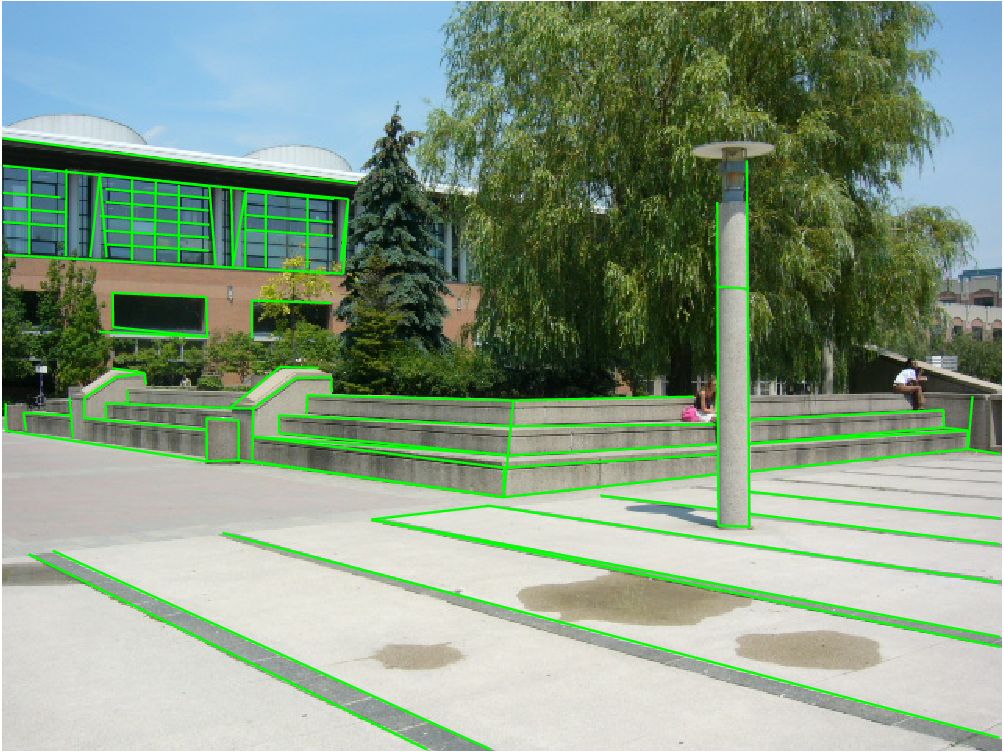} & 
    \includegraphics[width=0.3\textwidth]{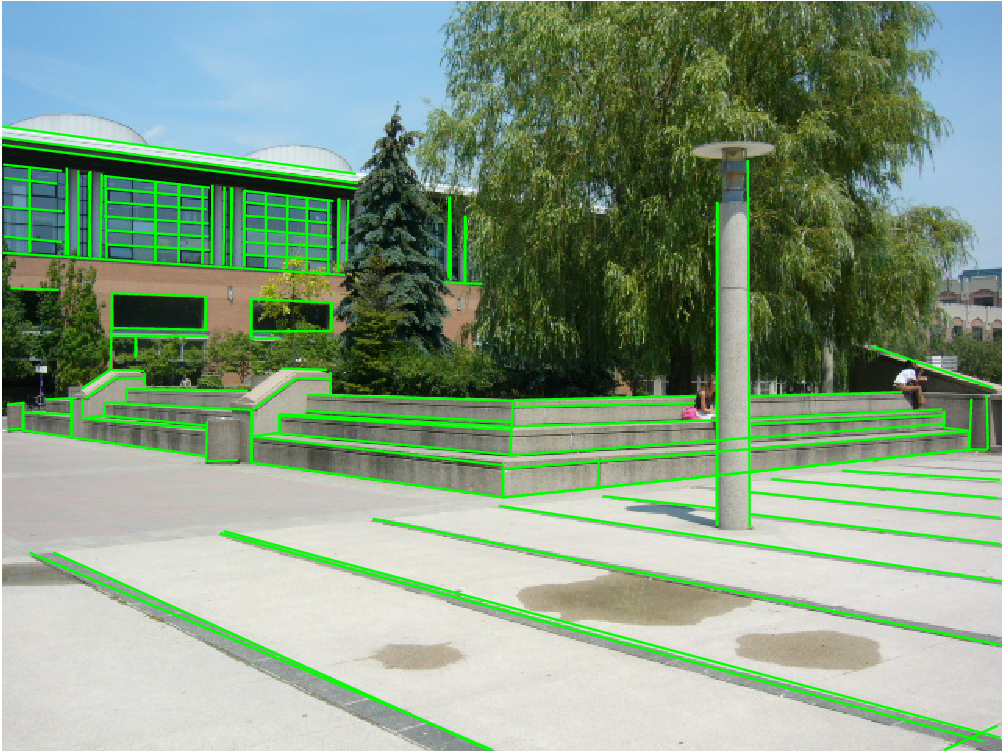} \\ 
(a) DeepLSD & (b) HAWPv2 & (c) DT-LSD \\
    \includegraphics[width=0.3\textwidth]{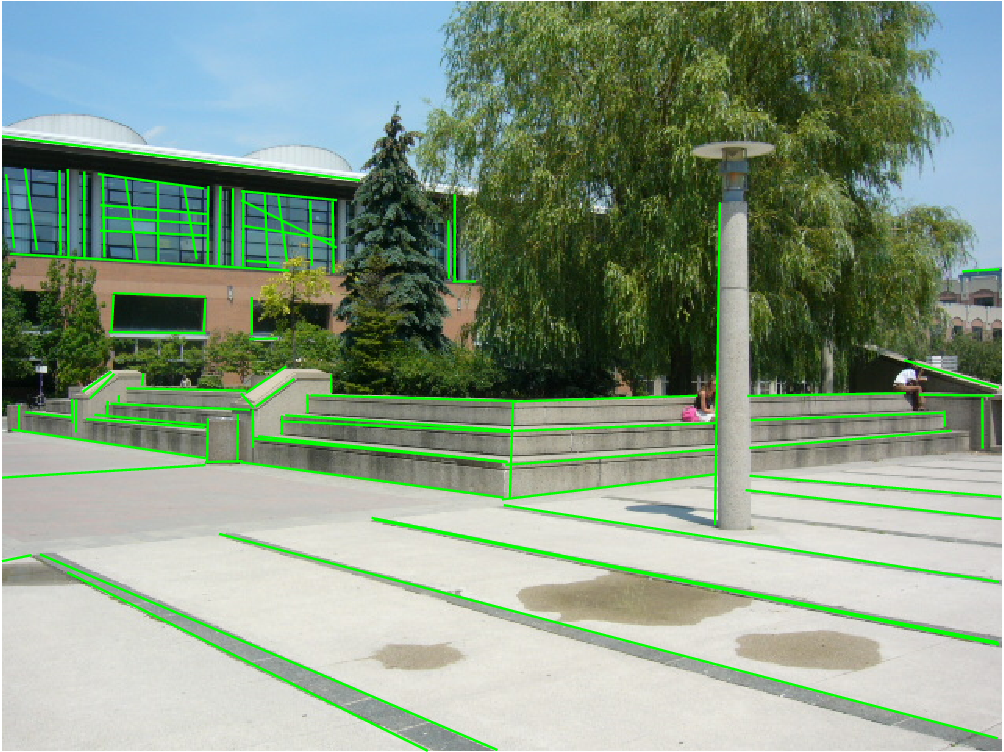} &
    \includegraphics[width=0.3\textwidth]{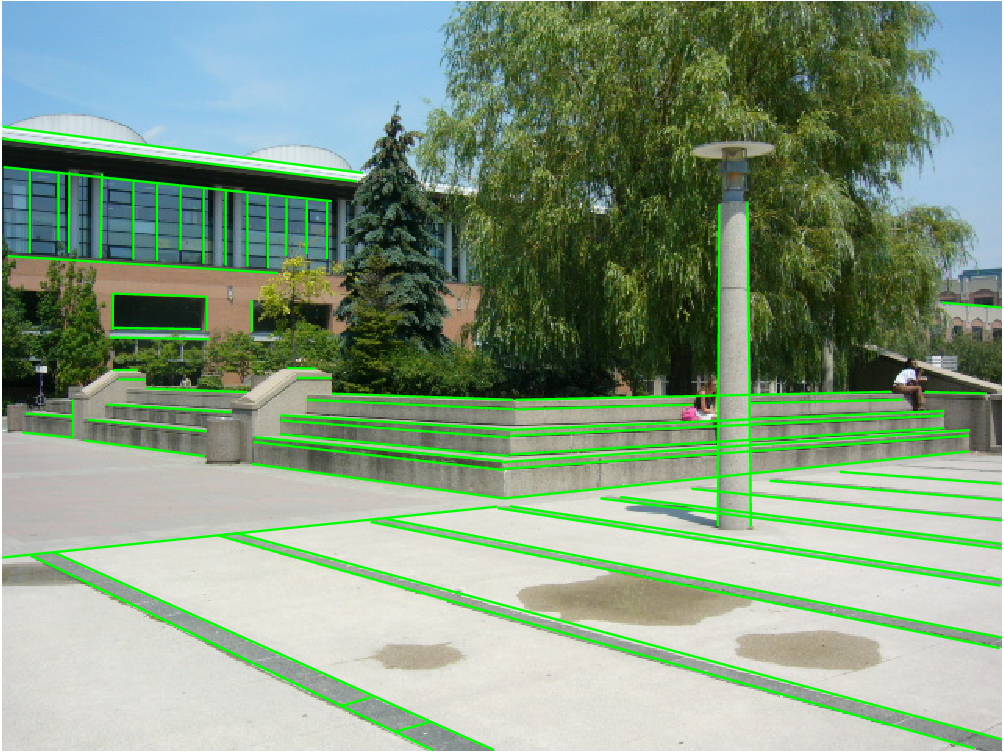} &
    \includegraphics[width=0.3\textwidth]{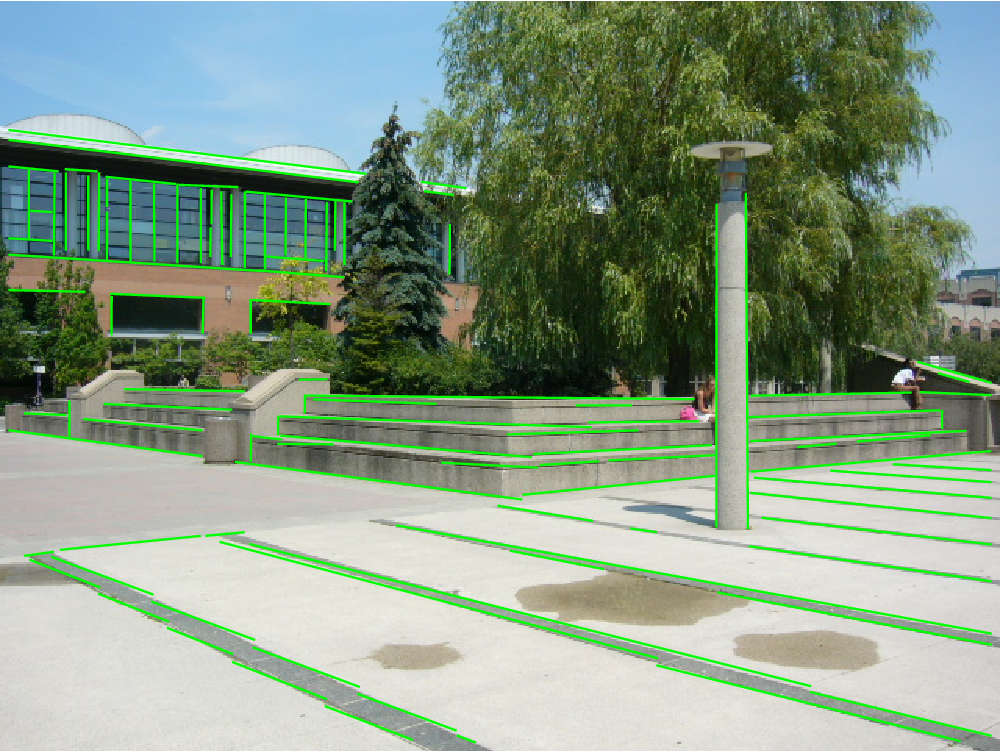} \\
(d) LINEA & (e) Ground truth  & (f) ADLA \\
\end{tabular}
  \caption{Line segment detection results computed by different methods on an image from the YorkUrban dataset}
  \label{york_image_original}
\end{figure}

Figure \ref{york_image_original} illustrates the same paradigm effect: 
many DeepLSD outputs are false positives under Manhattan annotations, 
whereas HAWPv2, DT-LSD, and LINEA suppress more non-Manhattan lines. 
ADLA is closest to the ground truth, 
retaining comparable true support with fewer false detections.

%% file: conclusions_revised.tex
\section{Conclusions}
\label{conclusions}

We introduced Adaptive Dual-Constrained Line Aggregation (ADLA) for generic, wireframe, and Manhattan line detection. 
ADLA separates task-appropriate edge evidence from a shared geometric extractor 
that combines orientation-coherence and orthogonal-distance constraints with progressive, edge-strength-weighted model refinement. 
Ablations show that the constraints improve accuracy, 
while adaptive refinement yields longer, less fragmented segments and further raises $\mathrm{F}^{\mathrm H}$.

ADLA achieves $\mathrm{F}^{\mathrm H}$ scores of 0.8665 on YorkUrban-LineSegment dataset, 0.8720 on ShanghaiTech dataset, and 0.7297 on YorkUrban dataset, 
outperforming the strongest baselines evaluated under the same protocol. 
Cross-paradigm comparisons confirm that rankings depend strongly on the annotation objective: 
wireframe-optimized methods do not necessarily generalize to exhaustive generic annotations. 
In contrast, 
ADLA remains strong across paradigms with limited parameter adjustment.

Performance is nevertheless bounded by the quality and task alignment of the input ESM. 
Moreover, 
ADLA does not explicitly model junction connectivity or global scene structure, 
so its segments can be less structurally complete than those of specialized wireframe detectors. 
Future work will add junction and connectivity reasoning while preserving cross-paradigm applicability and generic-line recall.

%% file: main.bbl
\begin{thebibliography}{10}
\expandafter\ifx\csname url\endcsname\relax
  \def\url#1{\texttt{#1}}\fi
\expandafter\ifx\csname urlprefix\endcsname\relax\def\urlprefix{URL }\fi
\expandafter\ifx\csname href\endcsname\relax
  \def\href#1#2{#2} \def\path#1{#1}\fi

\bibitem{vp2014}
J.~Lezama, R.~G. von Gioi, G.~Randall, J.-M. Morel, Finding vanishing points
  via point alignments in image primal and dual domains, in: 2014 IEEE
  Conference on Computer Vision and Pattern Recognition, 2014, pp. 509--515.

\bibitem{3D_line}
P.~Denis, J.~H. Elder, F.~J. Estrada, Efficient edge-based methods for
  estimating manhattan frames in urban imagery, in: European Conference on
  Computer Vision, 2008, pp. 197--210.

\bibitem{gpb}
P.~Arbelaez, M.~Maire, C.~Fowlkes, J.~Malik, Contour detection and hierarchical
  image segmentation, IEEE Transactions on Pattern Analysis and Machine
  Intelligence 33 (2011) 898--916.

\bibitem{object_line}
A.~Y.-S. Chia, D.~Rajan, M.~K. Leung, S.~Rahardja, Object recognition by
  discriminative combinations of line segments, ellipses, and appearance
  features, IEEE Transactions on Pattern Analysis and Machine Intelligence 34
  (2012) 1758--1772.

\bibitem{stereo_line}
Z.~Yu, X.~Guo, H.~Ling, A.~Lumsdaine, J.~Yu, Line assisted light field
  triangulation and stereo matching, in: 2013 IEEE International Conference on
  Computer Vision, 2013, pp. 2792--2799.

\bibitem{hough72}
R.~O. Duda, P.~E. Hart, Use of the hough transformation to detect lines and
  curves in pictures, Communications of the ACM 15 (1972) 11--15.

\bibitem{hough87}
J.~Skingley, A.~J. Rye, The hough transform applied to sar images for thin line
  detection, Pattern Recognition Letters 6 (1987) 61--67.

\bibitem{burns86}
J.~B. Burns, A.~R. Hanson, E.~M. Riseman, Extracting straight lines, IEEE
  Transactions on Pattern Analysis and Machine Intelligence PAMI-8~(4) (1986)
  425--455.

\bibitem{lsd}
R.~{Grompone von Gioi}, J.~Jakubowicz, J.-M. Morel, G.~Randall, Lsd: A fast
  line segment detector with a false detection control, IEEE Transactions on
  Pattern Analysis and Machine Intelligence 32 (2010) 722--732.

\bibitem{linelet}
N.-G. Cho, A.~Yuille, S.-W. Lee, A novel linelet-based representation for line
  segment detection, IEEE Transactions on Pattern Analysis and Machine
  Intelligence 40 (2018) 1195--1208.

\bibitem{meaningful00}
A.~Desolneux, L.~Moisan, J.-M. Morel, Meaningful alignments, International
  Journal of Computer Vision 40 (2000) 7--23.

\bibitem{dwp}
K.~Huang, Y.~Wang, Z.~Zhou, T.~Ding, S.~Gao, Y.~Ma, Learning to parse
  wireframes in images of man-made environments, in: The IEEE Conference on
  Computer Vision and Pattern Recognition (CVPR), 2018.

\bibitem{lcnn}
Y.~Zhou, H.~Qi, Y.~Ma, {End-to-End Wireframe Parsing}, in: IEEE Conference on
  Computer Vision (ICCV) 2019, 2019.

\bibitem{ppgnet}
Z.~Zhang, Z.~Li, N.~Bi, J.~Zheng, J.~Wang, K.~Huang, W.~Luo, Y.~Xu, S.~Gao,
  {PPGNet: Learning Point-Pair Graph for Line Segment Detection}, in: 2019
  IEEE/CVF Conference on Computer Vision and Pattern Recognition (CVPR), 2019.

\bibitem{afm}
N.~Xue, S.~Bai, F.~Wang, G.~Xia, T.~Wu, L.~Zhang, Learning attraction field
  representation for robust line segment detection, in: Proceedings of the IEEE
  Conference on Computer Vision and Pattern Recognition, 2019, pp. 1595--1603.

\bibitem{afmjournal}
N.~Xue, S.~Bai, F.-D. Wang, G.-S. Xia, T.~Wu, L.~Zhang, P.~H. Torr, Learning
  regional attraction for line segment detection, IEEE Transactions on Pattern
  Analysis and Machine Intelligence 43 (2021) 1998--2013.

\bibitem{hawp}
N.~Xue, T.~Wu, S.~Bai, F.-D. Wang, G.-S. Xia, L.~Zhang, P.~H.~S. Torr,
  Holistically-attracted wireframe parsing, in: IEEE Conference on Computer
  Vision and Pattern Recognition (CVPR), 2020.

\bibitem{hawpv2}
N.~Xue, T.~Wu, S.~Bai, F.-D. Wang, G.-S. Xia, L.~Zhang, P.~H.~S. Torr,
  Holistically-attracted wireframe parsing: From supervised to self-supervised
  learning, IEEE Transactions on Pattern Analysis and Machine Intelligence 45
  (2023) 14727--14744.

\bibitem{tplsd}
S.~Huang, F.~Qin, P.~Xiong, N.~Ding, Y.~He, X.~Liu, Tp-lsd: Tri-points based
  line segment detector, in: European Conference on Computer Vision, 2020, pp.
  770--785.

\bibitem{elsd}
H.~Zhang, Y.~Luo, F.~Qin, Y.~He, X.~Liu, Elsd: Efficient line segment detector
  and descriptor, in: 2021 IEEE/CVF International Conference on Computer Vision
  (ICCV), 2021, pp. 2949--2958.

\bibitem{dtlsd}
S.~Janampa, M.~Pattichis, Dt-lsd: Deformable transformer-based line segment
  detection, in: 2025 IEEE/CVF Winter Conference on Applications of Computer
  Vision (WACV), 2025, pp. 3477--3486.

\bibitem{linea}
S.~Janampa, M.~Pattichis, Linea: Fast and accurate line detection using
  scalable transformers, in: 2025 IEEE International Conference on Image
  Processing (ICIP), 2025, pp. 2115--2120.

\bibitem{rankletr}
X.~Tong, S.~Peng, B.~Tian, Y.~Guo, X.~Huang, Z.~Ma, Improving transformer based
  line segment detection with matched \\ predicting and re-ranking, in:
  Proceedings of the AAAI Conference on Artificial Intelligence, 2025.

\bibitem{canny}
J.~Canny, A computational approach to edge detection, IEEE Transactions on
  Pattern Analysis and Machine Intelligence PAMI-8 (1986) 679--698.

\bibitem{lsdsar}
C.~Liu, R.~Abergel, Y.~Gousseau, F.~Tupin, Lsdsar, a markovian a contrario
  framework for line segment detection in sar images, Pattern Recognition 98
  (2020).

\bibitem{ppht}
C.~Galamhos, J.~Matas, J.~Kittler, Progressive probabilistic hough transform
  for line detection, in: Proceedings. 1999 IEEE Computer Society Conference on
  Computer Vision and Pattern Recognition (Cat. No PR00149), Vol.~1, 1999, pp.
  554--560.

\bibitem{hough2011}
K.~Yang, S.~S. Ge, H.~He, Robust line detection using twoorthogonal direction
  image scanning, Computer Vision and Image Understanding 115 (2011)
  1207--1222.

\bibitem{hough2013}
D.~Shi, J.~Gao, P.~S. Rahmdel, M.~Antolovich, T.~Clark, Und: Unite-and-divide
  method in fourier and radon domains for line segment detection, IEEE
  Transactions on Image Processing 22 (2013) 2501--2506.

\bibitem{deeplsd}
R.~Pautrat, D.~Barath, V.~Larsson, M.~R. Oswald, M.~Pollefeys, Deeplsd: Line
  segment detection and refinement with deep image gradients, in: 2023 IEEE/CVF
  Conference on Computer Vision and Pattern Recognition (CVPR), 2023, pp.
  17327--17336.

\bibitem{aaglsd}
Z.~Li, A.~Shu, Aligned anchor groups guided line segment detector, in: Pattern
  Recognition and Computer Vision, 2026, pp. 3--16.

\bibitem{lgnn}
Q.~Meng, J.~Zhang, Q.~Hu, X.~He, J.~Yu, Lgnn: A context-aware line segment
  detector, in: Proceedings of the 28th ACM International Conference on
  Multimedia, 2020, pp. 4364--4372.

\bibitem{fclip}
X.~Dai, H.~Gong, S.~Wu, X.~Yuan, Y.~Ma, Fully convolutional line parsing,
  Neurocomputing 506 (2022) 1--11.

\bibitem{lsdnet}
L.~Teplyakov, L.~Erlygin, E.~Shvets, Lsdnet: Trainable modification of lsd
  algorithm for real-time line segment detection, IEEE Access 10 (2022)
  45256--45265.

\bibitem{mlsd}
G.~Gu, B.~Ko, S.~Go, S.-H. Lee, J.~Lee, M.~Shin, Towards light-weight and
  real-time line segment detection, in: Proceedings of the AAAI Conference on
  Artificial Intelligence, 2022.

\bibitem{emlsd}
S.~Hu, L.~Zhao, Q.~Wang, Em-lsd: A lightweight and efficient model for
  multi-scale line segment detection, Robotics and Autonomous Systems 195
  (2026) 105192.

\bibitem{deephough}
Y.~Lin, S.~L. Pintea, J.~C. van Gemert, Deep hough-transform line priors, in:
  European Conference on Computer Vision, 2020.

\bibitem{letr}
Y.~Xu, W.~Xu, D.~Cheung, Z.~Tu, Line segment detection using transformers
  without edges, in: 2021 IEEE/CVF Conference on Computer Vision and Pattern
  Recognition (CVPR), 2021, pp. 4255--4264.

\bibitem{detr}
N.~Carion, F.~Massa, G.~Synnaeve, N.~Usunier, A.~Kirillov, S.~Zagoruyko,
  End-to-end object detection with transformers, in: European Conference on
  Computer Vision, 2020, pp. 213--229.

\bibitem{scalelsd}
Z.~Ke, B.~Tan, X.~Zheng, Y.~Shen, T.~Wu, N.~Xue, Scalelsd: Scalable deep line
  segment detection streamlined, in: 2025 IEEE/CVF Conference on Computer
  Vision and Pattern Recognition (CVPR), 2025, pp. 6327--6336.

\bibitem{edgenat}
J.~Jie, Y.~Guo, G.~Wu, J.~Wu, B.~Hua, Edgenat: Transformer for efficient edge
  detection, in: ECAI 2024: 27th European Conference on Artificial
  Intelligence, 2024.

\bibitem{dinat}
A.~Hassani, H.~Shi, Dilated neighborhood attention transformer, preprint at
  \url{https://arxiv.org/abs/2209.15001v2} (2022).

\bibitem{dexined}
X.~Soria, A.~Sappa, P.~Humanante, A.~Akbarinia, Dense extreme inception network
  for edge detection, Pattern Recognition 139 (2023) 109461.

\bibitem{imagenet}
J.~Deng, W.~Dong, R.~Socher, L.-J. Li, K.~Li, F.-F. Li, Imagenet: A large-scale
  hierarchical image database, in: 2009 IEEE Conference on Computer Vision and
  Pattern Recognition, 2009, pp. 248--255.

\bibitem{structurededge}
P.~Dollar, C.~L. Zitnick, Fast edge detection using structured forests, IEEE
  Transactions on Pattern Analysis and Machine Intelligence 37 (2015)
  1558--1570.

\end{thebibliography}
